# Title: Prosody leaks into the memories of words

Kevin Tang[1]*, Jason A. Shaw[2]

**Affiliations:**

[1] Department of Linguistics, University of Florida, Gainesville, FL, U.S.A., 32611-5454. tang.kevin@ufl.edu

[2] Department of Linguistics, Yale University, New Haven, CT, U.S.A., 06520. jason.shaw@ufl.edu

*Correspondence to: tang.kevin@ufl.edu

**Running title**: prosody leaks

**ABSTRACT**

The average predictability (aka informativity) of a word in context has been shown to condition word duration (Seyfarth, 2014). All else being equal, words that tend to occur in more predictable environments are shorter than words that tend to occur in less predictable environments. One account of the informativity effect on duration is that the acoustic details of probabilistic reduction are stored as part of a word's mental representation. Other research has argued that predictability effects are tied to prosodic structure in integral ways. With the aim of assessing a potential prosodic basis for informativity effects in speech production, this study extends past work in two directions; it investigated informativity effects in another large language, Mandarin Chinese, and broadened the study beyond word duration to additional acoustic dimensions, pitch and intensity, known to index prosodic prominence. The acoustic information of content words was extracted from a large telephone conversation speech corpus with over 400,000 tokens and 6,000 word types spoken by 1,655 individuals and analyzed for the effect of informativity using frequency statistics estimated from a 431 million word subtitle corpus. Results indicated that words with low informativity have shorter durations, replicating the effect found in English. In addition, informativity had significant effects on maximum pitch and intensity, two phonetic dimensions related to prosodic prominence. Extending this interpretation, these results suggest that predictability is closely linked to prosodic prominence, and that the lexical representation of a word includes phonetic details associated with its average prosodic prominence in discourse. In other words, the lexicon absorbs prosodic influences on speech production.

**Declarations of interest**: none

**Funding:** This research did not receive any specific grant from funding agencies in the public, commercial, or not-for-profit sectors.

# 1 Introduction

The process of speech production leaves a phonetic imprint on words that reflects the particular communicative ecology of the speech act, including who's talking, who's listening, the common ground in the conversation, etc. If the product of the moment-by-moment influences on speech are stored in memory (i.e., lexicalized) and then accessed as targets for subsequent productions, word forms can drift in the direction of common influences on that word's production. In this way, words come to take on the phonetic characteristics of their typical usage contexts.

A key result in this vein is that the *average* contextual predictability of a word, i.e., the "informativity", predicts both word length (Piantadosi, Tily, & Gibson, 2011) and word duration (Seyfarth, 2014). This result builds on the well-established observation that word length and also word duration both vary with predictability. Frequently occurring words tend to have fewer segments than rarer words (Zipf, 1936). Even when the number of segments is held constant, word frequency negatively correlates with word duration, i.e., the millisecond duration of spoken words; frequent words tend to be shorter in duration than less frequent words (Bell et al., 2003; Jurafsky, Bell, Gregory, & Raymond, 2001). This type of probabilistic reduction of predictable words has been established in numerous languages, e.g., English (e.g., Bell, Brenier, Gregory, Girand, & Jurafsky, 2009), Dutch (Kuperman, Pluymaekers, Ernestus, & Baayen, 2007; Pluymaekers, Ernestus, & Harald Baayen, 2005a, 2005b), French (Bürki, Ernestus, Gendrot, Fougeron, & Frauenfelder, 2011; Pellegrino, Coupé, & Marsico, 2011; Torreira & Ernestus, 2011), Italian (Pellegrino et al., 2011), Spanish (Torreira & Ernestus, 2012), and Kaqchikel (Tang & Bennett, 2018). In contrast, the results showing effects of informativity (i.e., *average* predicatbility) on word duration are relatively new and are available only for English (Seyfarth, 2014, Sóskuthy & Hay, 2017).

That "on average" predictable words in English tend to be shorter than less predictable words, even in locally unpredictable contexts, may implicate phonetically detailed representations of words as well as a process of lexicalization (see, e.g., Bybee, 2001). A feedback loop, through which speaker productions are stored as phonetically detailed episodic memories, or "exemplars", can, over time, bias lexical representations in the direction of production constraints (e.g., Wedel, 2007).

There are two main aims of this paper. The first is to examine whether the effect of informativity on word duration extends to Mandarin Chinese, a typologically different language from English. Replicating this result in another language is important because of the implications that it has for the nature of lexical representations. The second aim is to evaluate whether the effect of informativity generalizes beyond word duration to other phonetic parameters, specifically pitch and intensity. This investigation is motivated by the hypothesis that phonetic reduction in predictable environments may reflect the structuring of language in terms of prominence and grouping, i.e., prosodic structure (Ladd 1986; Jun 2014).

Prosody is universal in the sense that all languages structure words into prosodic phrases of varying sizes and exhibit variation in word prominence. Prosodic structure has a substantial influence on the phonetic form of words, including their duration (e.g., Turk & Shattuck-Hufnagel, 2000). There is emerging evidence that the phonological properties of words are molded in part by the broader pragmatic goals of speech embodied in prosody (Roettger & Grice, 2019). Words produced at prosodic boundaries, under prosodic focus, or sentential stress are systematically enhanced, often produced with greater duration, intensity and pitch excursions than words in less prominent positions. The assignment of prosodic structure to speech varies according to a number of linguistic and para-linguistic factors and in language-specific ways. Notably, the perception of prosodic prominence is conditioned by the frequency of a word (Baumann 2014; Cole et al. 2010; Cole et al. 2017; Nenkova et al. 2007) and also by its informativity (Anttila et al., 2018).

## 1.1 Prosodic prominence and predictability

Results linking prosodic prominence to various measures of predictability provide a key motivation for the current study. Tradeoffs between predictability and phonetic robustness, i.e., how well the phonetic signal picks out the intended message, have been observed widely in natural language and have been hypothesized to have a uniform theoretical basis, rooted in the communicative function of human language (e.g., Jaeger 2010; Gibson et al., 2019; Hall et al., 2018). Moreover, prosody has been hypothesized to play a foundational role in this process (Aylett & Turk 2004; Turk 2010). According to the Smooth Signal Redundancy (SSR) hypothesis, prosody is the mechanism through which the speech signal is modulated to maintain high recognition probability (Turk 2010). Locally unpredictable words will therefore be enhanced, via increased prosodic prominence, in order to offset low language redundancy. Prosody affects not only temporal characteristics of speech, such as word duration, but also a number of other phonetic dimensions, including patterns of pitch variation and intensity, both of which contribute to word recognition probability in Mandarin Chinese (e.g., Ip & Cutler, 2020; Shaw & Tyler, 2020). We therefore explored whether effects of predictability found on duration in past studies would generalize beyond word duration to pitch and intensity.

## 1.2 Lexicalization of prosody

There is, by now, substantial evidence that memories for words encode both aspects of sound relevant to differentiating lexical items as well as aspects relevant to the circumstantial context in which a word was produced, including characteristics of the talker (Drager, 2011, Walker & Hay, 2011) and the physical location of the speech event (Hay et al., 2017). Moreover, it seems that circumstantial influences can accumulate over time within and across generations of speakers. The pronunciation of words co-evolves with usage patterns in a community such that changes in word frequency or typical prosodic position over time go hand in hand with changes in word duration (Sóskuthy & Hay, 2017). One possible account for such covariation relies on episodic memories of words. Consider a lexicon composed of exemplar clouds, i.e., clusters of episodic memories encoding the situation-specific phonetic details of a word (e.g., Pierrehumbert, 2001). On this view, any change in word usage with phonetic consequences, including the average prosodic prominence of the word, would naturally accumulate in the lexicon, shifting the phonetic distribution of the exemplar cloud in the direction of the usage-based influence. Since prosody is ubiquitous in natural speech, each word would come to reflect the typical prosodic environment in which it occurs.

## 1.3 Frequency, predictability, and informativity

It is in the context of the two lines of research summarized above that informativity, a measure of the average predictability of a word, becomes particularly relevant. Assuming that the assignment of prosodic structure, including prosodic prominence, is influenced in part by the local predictability of a word, following the Smooth Signal Redundancy hypothesis and empirical studies cited above, average predictability will reflect average prosodic prominence. Likewise, any lexicalized influence of prosodic prominence, e.g., pitch, intensity, duration, will be reflected in a word's average predictability. We therefore pursue informativity, defined as the average predictability of a word, as a key variable for assessing whether local influences of prosody, conditioned by local predictability, have accrued in the lexicon.

Notably, informativity, as defined here, is a word-level variable. In this way, informativity is like word frequency and it differs from local predictability, which is not a property of words but a property of tokens, i.e., instances of a word in a particular usage context. As sketched above, each of these variables, frequency, predictability and informativity, has a distinct theoretical interpretation: word frequency as the size of a word's exemplar cloud; predictability as local prosodic prominence; informativity as average prosodic prominence. However, while they reference distinct theoretical constructs, the three variables are likely to be correlated—frequent words are likely to be locally predictable, and, therefore, have low informativity. Fortunately, effects of informativity can be reliably differentiated from frequency/predictability, even in the presence of natural interdependency between these variables, as long as all three are factored into the analysis (for a demonstration through computational simulation, see Cohen Priva & Jaeger, 2018).

*1.4 Mechanisms for robust communication*

There is much debate about the cognitive mechanism responsible for the observation that word length/duration correlates with contextual predictability (see, Jaeger and Buz (2017) for an overview). On one account, the correlation follows from the architecture of the speech production system, such that speech rate slows according to factors that slow lexical access in production (e.g., Bell et al., 2009). Another account is that speakers actively balance contextual predictability with signal robustness (e.g., Jaeger, 2015; Wedel, Nelson, & Sharp, 2018) possibly with audience design in mind (Watson, Arnold, & Tanenhaus, 2008, 2010). Fewer resources in production are expended when communication is not at risk (predictable contexts); additional production resources are drawn upon in challenging communication environments (unpredictable contexts, noisy environments, etc.). The hypothesis that prosody functions to highlight unpredictable words serves to focus this account. Specifically, it predicts that the resources of prosodic modulation are the resources deployed to enhance signal redundancy.

To test our hypotheses, in the current study we expand beyond word duration to also investigate pitch and intensity, dimensions known to encode prosodic prominence in Mandarin Chinese. Following antecedent research, we investigate the effects of both forward predictability and backward predictability, as the directionality of the predictability effect may also be revealing as to the cognitive mechanism responsible for it.[1] To test whether influences of local predictability on phonetic measures of prosodic prominence have been lexicalized, we additionally factor informativity into our analysis. If informativity is picking up on lexicalized prosodic prominence, then informativity should predict not just word duration but also other phonetic reflexes of prosodic prominence, e.g., pitch, intensity. We test this prediction in a series of studies on Mandarin Chinese.

In what follows, we describe four corpus studies of Mandarin Chinese. In each case, we fit linear mixed effects models to phonetic parameters related to prosodic prominence: duration, pitch, and intensity. Through a combination of fixed and random effects, we investigate the influence of frequency, contextual predictability and informativity while controlling for as many additional factors known to influence these variables as possible. The first three studies investigate word duration, pitch and intensity, respectively. The fourth study is a mediation analysis, investigating whether effects of informativity on pitch and intensity are mediated by duration.

---

[1] The definition of predictability is provided in the methods section (2.2.1). Forward predictability refers to the probability of a word given the preceding word; backward predictability refers to the probability of a word given the following word. Sometimes these quantities are described as preceding/following predictability (e.g., Bell et al. 2009) instead of forward/backward predictability (e.g., Tremblay & Tucker 2011).

## 2 Materials and Methods

*2.1 Acoustic corpus*

Acoustic measurements were extracted from the HKUST Mandarin Telephone Speech, Part 1 corpus developed by Hong Kong University of Science and Technology (HKUST) (Fung, Huang, & Graff, 2005). The corpus is a collection of 150 hours of Mandarin Chinese conversational telephone speech from Mandarin speakers in mainland China. 1,793 speakers were recruited from several cities across mainland China. Most of the speakers did not know each other. Two speakers were connected by a telephone operator and they were assigned a specific topic from 40 topics to encourage a more meaningful conversation. Each call was capped at 10 minutes and the majority of the calls reached this limit. All but one speaker spoke only in one call. In total, 897 ten-minute long calls, each with two speakers having a conversation on an assigned topic, were recorded. Each side of a call was recorded in two separate files. Some demographic information of the speakers was available, such as age, gender and phone type (a fixed landline connection or a mobile connection).

The corpus contains the audio recordings and their corresponding orthographic transcriptions using Chinese characters with utterance-level timestamps. In addition, the transcriptions contain a range of annotations concerning disfluent speech (e.g., partial words, restarts, filled pause), speaker noise (e.g., laughers, coughs), background noise, hard to understand speech regions, and use of foreign (non-Chinese) languages. For further information concerning the annotations of the corpus, see the transcription guidelines for EARS Chinese telephony conversational speech database (https://catalog.ldc.upenn.edu/docs/LDC2005S15/trans-guidelines.pdf).

To obtain the acoustic measurements, the corpus needed to be forced aligned. At the time of writing, the authors were not aware of any available forced aligner for Mandarin Chinese, with the exception of SPPAS (Bigi, 2015), since the alignment quality of SPPAS was unacceptable, we therefore decided to train our own aligner using the Montreal Forced Aligner (MFA, version 1.0) (McAuliffe, Socolof, Mihuc, Wagner, & Sonderegger, 2017). For details of the forced alignment process, including corpus segmentation, data filtering, word segmentation, and part of speech tagging, see supplementary materials.

*2.1.6 Acoustic estimates*

Using the word alignment, three acoustic dimensions of each word were estimated: duration, intensity (maximum, minimum) and pitch (maximum, minimum). Duration was extracted directly from the textgrids. Intensity and pitch measurements were made using Praat (Boersma & Weenink, 2019). An intensity object was created using the function 'To Intensity…' with the default parameters (Minimum pitch was set as 100.0 Hz and the time step was computed as one quarter of the effective window length). A pitch object was created using the function 'To Pitch…' with the default parameters (pitch floor was set to 75 Hz, pitch ceiling was set to 600 Hz and time step was 0.75 / pitch floor). The maximum and minimum intensity and pitch were extracted after a parabolic interpolation. For each word, we obtained five acoustic variables: duration, maximum intensity, intensity range (maximum intensity minus minimum intensity), maximum pitch and pitch range (maximum pitch minus minimum pitch).

*2.2 Lexical corpus*

In order to examine the effect of word probabilities, a speech-like written corpus of Mandarin Chinese (Tang & Mandera, In preparation) was used to estimate word frequency, contextual predictability and informativity. The corpus consists of 431 million word tokens from TV/film subtitle texts of Mandarin Chinese. The written corpus was word segmented and POS tagged using ICTCLAS just as the transcriptions of the acoustic corpus. This corpus was chosen for a number of reasons.

Firstly, previous work has shown that frequency estimates derived from subtitle texts consistently outperform those from non-speech like genres (such as newspaper texts) in explaining behavioural data, such as reaction time in lexical decision tasks and word naming tasks (Brysbaert & Boris New, 2009; Cai & Brysbaert, 2010; Keuleers, Brysbaert, & Boris New, 2010; van Heuven, Mandera, Keuleers, & Brysbaert, 2014). This is even the case when the non-speech-like corpora are larger in size than the corresponding subtitle corpus.

Secondly, subtitle texts are a better genre-of-speech match with the telephone conversational speech compared to newspaper texts in terms of their levels of formality and the fact that subtitle texts consist primarily of dialogues.

Thirdly, its large corpus size improves the representativeness of the corpus (Biber, 1993), under the typical assumption that the larger your corpus the more representative it is as the average linguistic experience of the language users.

Finally and most importantly, informativity computed using a large corpus will reduce the chance of frequency falsely capturing effects that should be attributed to informativity. In a simulation study, Cohen Priva and Jaeger (2018) investigated whether the size of the lexical corpus could create spurious frequency and predictability effects. First, they created lexical corpora of various sizes sampled from a bigger corpus which they assumed to represent the "true" experience of the language users. Second, they computed frequency, contextual predictability and informativity using each of these sampled corpora as well as the "true" corpus. Finally, they correlated the three estimated predictability variables computed using the sub corpora with each of the predictability variables computed using the "true" corpus. They reported that even for a sampled lexical corpus of 10 million words, they found the estimated frequency variable would still spuriously correlate with the "true" informativity variable. While the study did not report the minimum lexical corpus size required to avoid or mediate a spurious frequency effect, it is nonetheless clear that our corpus of over 400 million words is a substantial methodological improvement compared to the 12 million word corpus used by other studies on word informativity (Seyfarth, 2014). This is particularly true, considering that a) Mandarin Chinese has relatively simple morphology compared to English, so the inter-word probability estimates (bigrams) should be even better in Mandarin Chinese, and b) the 12 million word corpus in English was already sufficient to reveal an informativity effect.

*2.2.1 Probability estimates*

The three word probability measures are described below.

**Word frequency** is the total number of times a word appears in the lexical corpus.

**Contextual predictability** is the conditional probability of a word given its context as formulated in equation 1, where $c$ is the context which is operationalized as the preceding or following word in an utterance and $w$ is the target word.

$$\text{Equation 1: } Pr(W = w \mid C = c)$$

**Informativity** is the negative log average contextual predictability of a word in every context in which it appears in, weighted by the contextual predictability of the contexts given the word, as formulated in equation 2.

$$\text{Equation 2: } -\sum_c Pr(C = c \mid W = w) \log_2 Pr(W = w \mid C = c)$$

Following Seyfarth (2014), contextual predictability was computed using language models. Two bigram language models were constructed using the lexical corpus. One model describes the probability of each word given the word before it (the previous word), and the other model describes the probability of each word given the word after it (the following word). Model construction was carried out using the MIT Language Modeling (MITLM) toolkit (Hsu, 2009). The probabilities in the language models were smoothed using the modified Kneser-Ney method (Chen and Goodman, 1999) with the default smoothing parameters provided by the toolkit. These two models were used to estimate the contextual predictability and the informativity of each word in the spoken corpus. The language models were fitted over word types that occur more than once; this ensures that the models can be fitted and it reduced the total number of word types from over 900,000 to 470,000. The informativity measures were computed over bigrams that occur more than once, also to ensure computationally feasibility. This reduced the total bigram types from 25 million to 9.2 million.

Figure 1 illustrates the relationship between word frequency and informativity. Two lexical items with similar frequency but different informativity are highlighted in the figure.

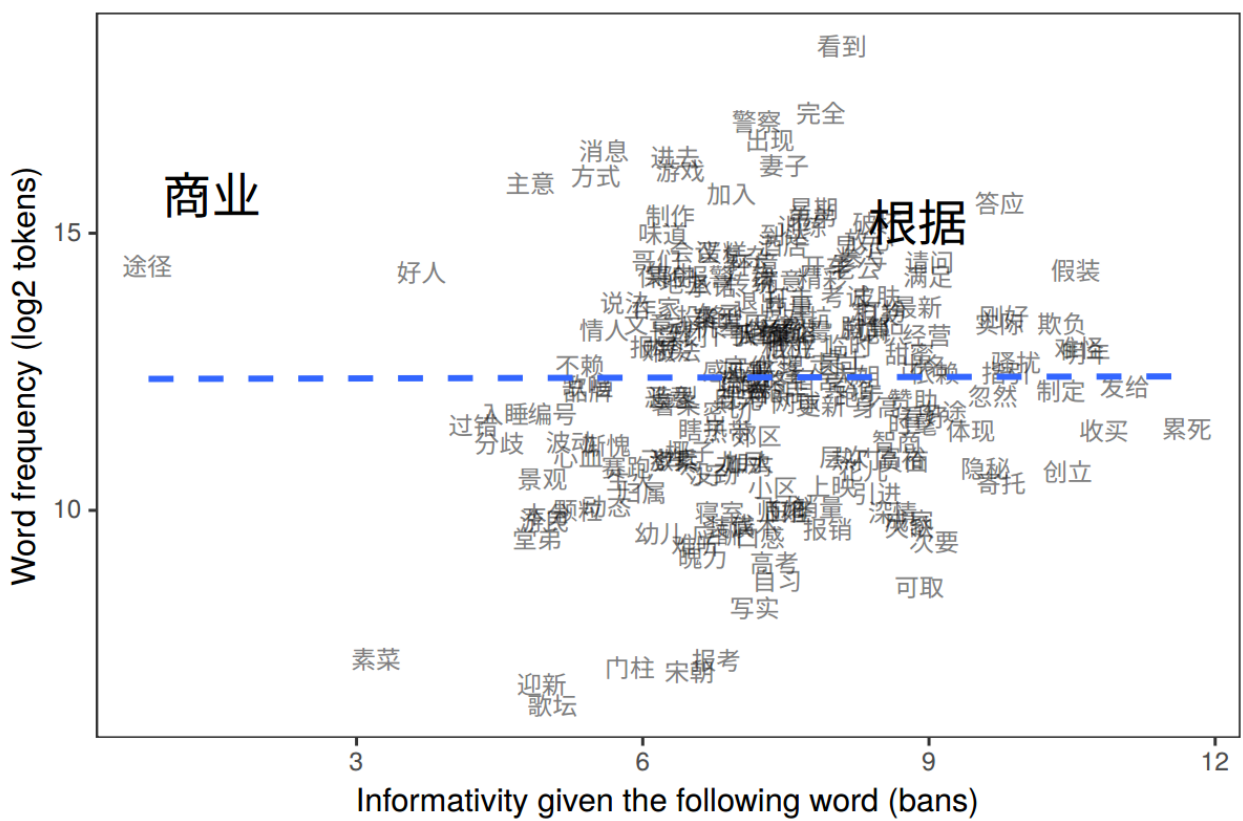

Figure 1: An illustration of the relation between frequency and informativity given the following word in Mandarin Chinese, modelled after Seyfarth (2014)'s illustration on English words. The figure shows a sample of 200 word types that were observed in the acoustic data at least 10 times. The dotted line indicates the trend between frequency and informativity for the sampled words. Two specific words with similar frequency but different informativity (商业 'business' and 根据 'according to') are highlighted using a larger font size.

*2.3 Variables*

Words in the acoustic corpus were annotated based on the lexical corpus and the acoustic corpus for a range of type-level and token-level variables. They are described below for dependent variables and predictor variables separately. Besides the informativity variables, the rest of the predictor variables were reported to have an effect on duration in previous literature (e.g., Seyfarth 2014; Gahl 2008). To control for inherent prosody of the segmental and tonal composition of the words, three baseline variables for duration, pitch and intensity were computed as described below.

*2.3.1 Dependent variables*

**Duration:** word duration (in ms) was extracted using the word-level timestamps. It is a continuous variable and log-transformed (base 10).

**Maximum intensity:** maximum intensity (in dB) was extracted from the intensity contour of each word. It is a continuous variable. No log-transformation was needed because the decibel is already on a log scale.

**Intensity range:** intensity range (in dB) was computed using the maximum intensity and minimum intensity measurements. It is a continuous variable with no additional log-transformation.

**Maximum pitch:** maximum pitch (in Hz) was extracted from the pitch contour of each word. It is a continuous variable and log-transformed (base 10).

**Pitch range:** pitch range (in Hz) was computed using the maximum pitch and minimum pitch measurements. It is a continuous variable and log-transformed (base 10).

*2.3.2 Baseline variables*

A syllable level baseline of each of the dependent variables (duration, maximum pitch and maximum intensity) was estimated using pointwise values as described in the following steps. The role of the baseline is to incorporate the contribution of segments and tones to the dependent measures of interest.

1) The duration, pitch and intensity values of all syllables were extracted for each spoken word token.
2) The extracted syllable-level values were by-speaker normalized into z-scores to capture speaker-level information. The mean and standard deviation values of each speaker and each acoustic dimension were stored for Step 4).
3) For every syllable token, all z-normalized values of the same syllable type, except the value of the target token, were averaged. The purpose of excluding the value of the target syllable token from the averaging process was to create a pointwise acoustic estimate of that specific syllable token.
4) The z-normalized pointwise averaged acoustic estimates were converted back to a real acoustic space by using the mean and the standard deviation of the actual acoustic values for each speaker separately.
5) For monosyllabic words, the pointwise averaged syllable-level value was used as the word-level baseline value for word duration, pitch and intensity. Polysyllabic words were estimated differently. For word duration, the baseline was the sum of the pointwise averaged syllable-level duration values of each syllable in the word. For max pitch and max intensity, the highest of the pointwise averaged syllable-level pitch/intensity values was used as the word-level maximum pitch/intensity baseline.

*2.3.3 Predictor variables*

**Token frequency:** the number of times a word appears in the lexical corpus were counted, log-transformed (base 2) and then z-transformed.

**Contextual predictability:** two variables of contextual predictability were estimated from the lexical corpus using equation 1: a) the conditional probability of a word given the previous word (*forward predictability*) and b) the conditional probability of a word given the following word (*backward predictability*); both were log-transformed (base 2) and then z-transformed.

**Informativity:** two variables of informativity were estimated using the lexical corpus according to equation 2: a) the informativity of a word given the previous word (*forward informativity*) and b) the informativity of a word given the following word (*backward informativity*); both were log-transformed (base 2) and then z-transformed.

**Word length:** the number of segments in the word transcription was counted and then z-transformed. Word length serves as a partial control for word duration, the more segments a word has, the longer its duration. The effect of word length on intensity and pitch is less clear. It is possible that longer words are more likely to have high intensity/pitch as there would be more time to achieve high intensity and pitch targets.

**Disfluency:** two binary variables of disfluency were estimated using the annotation of the acoustic corpus, *preceding disfluency* and *following disfluency*. The variables indicate whether the word is immediately a) preceded and b) followed by a non-silence disfluency, namely laughters, sneezes, coughs, lipsmacks and filled pauses; these were sum-coded with 'not disfluent' being the reference level.

**Pause duration:** two continuous variables of pause duration within an utterance were estimated for the alignment of the acoustic corpus using short pauses detected automatically by the trained aligner as well as the duration of annotated-then-aligned breath units: *preceding pause duration* and f*ollowing pause duration*. The variables are the duration of a pause immediately a) preceding the word and b) following a word respectively. The duration variables (in ms) were Laplace transformed (add one), log-transformed (base 10) and then z-transformed.

Pause duration is used as an approximation to determining phrasal position and boundary strength. Previous work on word duration found that words before a pause have longer word duration, which suggests phrase-final lengthening. Rather than coding pauses as a binary variable, as was done, for example, in Gahl (2008) with an arbitrary cut-off duration of 0.5 second, we coded it as a gradient variable to provide a more accurate estimate for phrasal positional effects, since boundary strength as estimated with pause duration has been shown to predict the rate of segment deletion (Tanner, Sonderegger, & Wagner, 2017). Following the practice of Tanner et al. (2017), force-aligned pauses of less than 30 ms were set to have a duration of 0 ms because they are likely to be aligner errors or due to low amplitude signals (such as stop closures).

**Speech rate:** Speech rate was estimated as the number of syllables per second (de Jong & Wempe, 2009). Following the practice of Gahl (2008) and Seyfarth (2014), for the purpose of computing speech rate, an utterance is defined as a stretch of speech within a conversational turn (which has a maximum duration of 10 seconds as defined by the corpus developers) that are marked by pauses, disfluencies, and other interruptions that are longer or equal to 0.5 seconds or by the conversational turn boundaries. Two continuous variables of speech rate were computed, *preceding speech rate* and *following speech rate*. They are the speech rate on either a) the left or b) the right of the target word.

They were computed using the number of syllables before/after the target word itself, divided by the duration of that region, and then z-transformed.

**Previous mention**: Previous work (Fowler, 1988; Fowler & Housum, 1987) has shown that words which are repeated in a spoken discourse are sometimes reduced in production compared to previous mentions of those words. Repetitions were coded separately for the previous mention of a word from the same speaker and those from another speaker within the dialogue, since it has been shown that these two types of repetitions can have different effect sizes on acoustic reduction (Tron, 2008). Two binary variables were computed, *self-mention* and *cross-speaker mention*; sum-coded with 'no previous mention' as the reference level[2].

**Syntactic category:** the syntactic category of the words was coded using the main tags in the ICTPOS 3.0 tag set. After excluding certain categories (as outlined in Section 2.4), nine categories remained. This variable was coded using the target encoding (also called mean encoding) scheme (Micci-Barreca, 2001), which takes the mean of the dependent variable for each category to yield a single continuous variable. This variable was then z-transformed. The target encoding scheme was chosen over the usual contrast coding schemes, because, firstly it greatly reduces the number of predictors needed to code a nine-level categorical variable from eight predictors (N-1) to just one; secondly, it does not sacrifice any details of the nine categories; and finally, it performs similarly to or better than contrast coding schemes in regression and classification models (Cerda, Varoquaux, & Kégl, 2018).

**Age**: the age of the speaker was included to capture potential social factors. The age variable was in years and z-transformed.

**Gender**: similar to age, the gender of the speaker was included to capture potential social factors. The gender variable was binary and sum-coded with 'Female' as the reference level.

*2.4 Exclusion criteria*

After the acoustic estimates and the probability estimates were computed, certain acoustic word tokens were excluded given a number of criteria as outlined below. The final dataset consisted of 417,756 words.

a) Words for which we could not compute all of the acoustic estimates (44,875 words). Specifically, Praat failed to compute the intensity and pitch values in these cases.
b) Words that have impossible acoustic values such as a negative intensity values[3] (7 words).
c) Words that cannot be part-of-speech tagged by the ICTCLAS tagger (755 words).
d) Words that are tagged as being proper names (22,261 words) and other miscellaneous tags (6,270 words) such as onomatopoeia. Proper names have been shown to behave differently from typical nouns in that their probabilistic estimates are speaker dependent (Bredart, 2002; Cohen & Burke, 1993; Nomi & Cleary, 2008).
e) Words that are tagged as function words such as pronouns, classifiers, prepositions and others (537,815 words). Function words were not analysed in this study, since it has been shown that predictability has different effects on the duration of function words and content words (Bell et al., 2009; Tang & Bennett, 2018).
f) Words that are annotated as being produced only partially (15,370 words) or mispronounced (654 words). These words were excluded because their acoustic details are shown to differ from typically produced words. Partially produced words have shorter segmental content and acoustic details that differ from their fully produced forms (Howell & Vause, 1986; Howell & Williams, 1992). Words that are mispronounced have shown to have both categorical and gradient acoustic errors (Frisch & Wright, 2002; Goldrick, Keshet, Gustafson, Heller, & Needle, 2016).
g) Words that are filled pause words (23,302 words), acronyms (2,056 words) or foreign words (467 words).
h) Words that appear in the corpus only once. This improves the interpretation of the statistical models, since our models have random intercepts of word types and random effects, which are best used for repeated levels.
i) Words that are at the start or at the end of a conversational turn (as previously mentioned in Section 2.1.1 Corpus segmentation), since contextual predictability and speech rate cannot be computed.
j) Words that are at the start or at the end of an utterance (as previously defined in Section 2.3.1 Dependent variables for computing speech rate), since speech rate cannot be computed.
k) Words for which we cannot compute pointwise acoustic baseline measures.

---

[2] Cross-speaker-mention was computed differently from self-mention. Some conversational turns were not aligned due to their inherent noise or the speaker's phone type (see supplementary materials for details). We, therefore, could not always use the word-level timestamps to check if a word produced by one speaker was previously mentioned by the other speaker. We opted for a conservative coding scheme, such that a word by the other speaker is only counted as mentioned if its conversational-turn-level offset timestamp comes before the word-level onset timestamp of the target word by the speaker. Note that this means some cross-speaker previous mentions could have been missed in cases of cross-talk.

[3] Manual inspections revealed that these negative intensity values are due to the signal consists of mainly silence.

## 2.5 Model procedure

Linear mixed-effects models were fit to the acoustic data conducted using the MixedModels (v3.0.0) package in Julia (Version 1.5.2) (Bates et al., 2020; Bezanson et al., 2017). For each of the five dependent variables (one duration, two intensity, and two pitch variables), a model was fitted with the predictor variables outlined in Section 2.3.3 as fixed effects. Furthermore, tones are known to condition syllable duration (Yang, Zhang, Li, & Xu, 2017) and intensity (Liu & Samuel, 2004; Whalen & Xu, 1992) and, by definition, pitch. To capture the inherent prosody of the words, a baseline variable outlined in Section 2.3.2 was included as a fixed effect. These models were fitted with a number of random effects, to which we now turn.

As is typical of psycholinguistic research, per-word random intercepts and per-speaker random intercepts were included to allow for idiosyncrasies of individual speakers (1,655 speakers) and word types (6,347 word types). In addition to these random intercepts, three correlated per-speaker slopes of frequency, backward informativity and forward informativity were fitted to allow for the frequency and informativity effects to vary by speaker. P-values for each effect were calculated using the normal approximation to the *t*-statistic. While it is not as ideal as using the Satterthwaite approximation (Luke, 2017)[4], given our large sample size (>400,000 tokens), the p-values should not be particularly anti-conservative (Barr, Levy, Scheepers, & Tily, 2013).

---

[4] Owing to the size of our models, we were unable to compute the Satterthwaite approximation (Luke, 2017) to get approximate degrees of freedom as implemented in the *lmerTest* library (Kuznetsova, Brockhoff, & Christensen, 2017).

The structure of the models is given below in the syntax of *lmer*.

*Dependent variable (either Duration, Maximum intensity, Intensity range, Maximum pitch or Pitch range) ~ Pointwise syllable-based baseline (either Duration, Maximum intensity or Maximum Pitch)[5] + Frequency + Forward predictability + Backward predictability + Forward informativity + Backward informativity + Word length + Preceding disfluency + Following disfluency + Preceding pause duration + Following pause duration + Preceding speech rate + Following speech rate + Previous self-mention + Previous cross-speaker mention + Age + Gender + Syntactic category + (1 | Word type) + (1 + Frequency + Forward informativity + Backward informativity | Speaker)*

In addition to these models, a series of mediation analyses was conducted to examine whether the effects of informativity, if any, found in one model for a dependent variable can be explained by another dependent variable. This was particularly important for the two acoustic dimensions, pitch and intensity, that were not previously examined for informativity effects. Should there be an informativity effect found for duration and also for pitch and/or intensity, we would need to rule out duration being a mediator that underlies the observed relationship between pitch/intensity and informativity. For completeness, such mediation analyses were also conducted for duration with pitch and/or intensity being the mediator(s). These analyses were done by adding the mediator variable as an additional fixed effect.

All models underwent the process of model criticism. For each model, the residuals were extracted and data points that were 2.5 standard deviations above or below the mean residual value were excluded. No more than 4% of the data points were excluded in any of the models.[6] To evaluate potential collinearity issues, we computed the Variance Inflation Factor (VIF) of the predictor variables in each of the models. In all cases, all variables have VIF < 10 which indicates no serious issues of collinearity.

Table 1 summarizes the distribution of the variables (both dependent variables and the predictors). The tables show the mean, standard deviation, interquartile range and range (max-min) for the continuous variables and count information for the categorical variables.

---

[5]The models for pitch range and intensity range used the same baseline as maximum pitch and maximum intensity respectively. Range baselines would only be possible for polysyllabic words, since a monosyllabic word would only have one pointwise averaged value, while a polysyllabic word would have multiple pointwise averaged values and the range baseline would be the difference between the maximum and the minimum pointwise averaged syllable-level values. We therefore opted to use the same baseline for maximum pitch and pitch range and similarly for maximum intensity and intensity range to maintain consistency of the baseline used across word sizes,

[6] This strategy is a potential way to deal with non-normality or overly influential observations. As suggested by our reviewers, some statisticians do not recommend this (Harrell, 2015). Given the different philosophical practices, we fitted our models with and without trimming . In all the models that were evaluated, the effect of trimming did not make a qualitative difference to predictability variables and it generally increased the level of statistical significance .

|  | Mean | SD | IQR | Range |
|---|---|---|---|---|
| Word duration (log10, ms) | 2.3252 | 0.2174 | 0.3153 | 1.4244 |
| Maximum Pitch (log10, hz) | 2.3367 | 0.1713 | 0.2525 | 0.9320 |
| Maximum Intensity (dB) | 70.8248 | 8.1746 | 11.2467 | 64.8950 |
| Pitch range (log10, hz) | 1.4943 | 0.4933 | 0.6136 | 2.7344 |
| Intensity range (dB) | 18.1750 | 11.1220 | 16.9216 | 61.6008 |
| Duration baseline (log10, ms) | 2.3585 | 0.1702 | 0.2886 | 1.3689 |
| Pitch baseline (log10, hz) | 2.3004 | 0.1164 | 0.1989 | 0.8509 |
| Intensity baseline (dB) | 66.9850 | 6.8223 | 9.3866 | 50.8231 |
| Frequency (log2) | 18.6384 | 3.3388 | 5.1273 | 20.6272 |
| Forward predictability (log2) | -8.7663 | 3.8368 | 5.4807 | 24.0381 |
| Backward predictability (log2) | -8.8207 | 4.0109 | 6.0244 | 24.2576 |
| Forward informativity (log2) | 6.6645 | 2.0969 | 3.2477 | 15.6565 |
| Backward informativity (log2) | 6.2777 | 1.6385 | 2.4591 | 14.3945 |
| Word length | 3.1133 | 1.3312 | 2.0000 | 11.0000 |
| Preceding pause duration (log10, ms) | 0.1994 | 0.6325 | 0.0000 | 2.6911 |
| Following pause duration (log10, ms) | 0.2014 | 0.6570 | 0.0000 | 3.4844 |
| Preceding speech rate | 5.5856 | 1.8898 | 2.3188 | 32.9710 |
| Following speech rate | 5.2407 | 1.6838 | 2.0075 | 33.0929 |

| Age | 27.4256 | 4.7909 | 6.0000 | 41.0000 |
|---|---|---|---|---|
| Syntactic category (target coded to word duration) | 2.3252 | 0.0895 | 0.0739 | 0.3582 |
| Syntactic category (target coded to maximum pitch) | 2.3367 | 0.0100 | 0.0127 | 0.0699 |
| Syntactic category (target coded to maximum intensity) | 70.8248 | 1.2018 | 0.7893 | 4.9330 |
| Syntactic category (target coded to pitch range) | 1.4943 | 0.0943 | 0.1378 | 0.4562 |
| Syntactic category (target coded to intensity range) | 18.1750 | 2.0269 | 3.8860 | 11.0776 |
| Preceding disfluency | True: 5,239, False: 412,517 | | | |
| Following disfluency | True: 2,475, False: 415,281 | | | |
| Previous self-mention | True: 270,302, False: 147,454 | | | |
| Previous cross-speaker mention | True: 230,480, False: 187,276 | | | |
| Gender | Male: 210,169, Female: 207,587 | | | |

Table 1: *Descriptive statistics of variables*

## 3 Results

Before examining the effect of informativity on acoustic prominence, we conducted a correlation analysis between the five acoustic variables (after transformation) as shown in table 2. This will allow us to better understand if any informativity effect found for a given acoustic variable is likely to be a spillover effect from another correlated acoustic variable.

All correlations were statistically significant due to the large amount of data. Duration was most strongly correlated with intensity range (R=0.56) and pitch range (R=0.50), c.f., maximum intensity (R=0.23) and max pitch (R=0.15). Maximum pitch and maximum intensity were weakly correlated at 0.12. These correlational relationships suggest that the effect of informativity on prominence would be most conclusive if the informativity effect were found with duration, maximum intensity and maximum pitch, because they are the least correlated with each other. For this reason, we focus on these three dependent measures.

Four studies were conducted. Study I focuses on the effect of informativity on word duration, replicating Seyfarth (2014) in a new language, Mandarin Chinese. Study II extends the effect to pitch (maximum pitch) which is another phonetic cue to prosodic prominence in Mandarin Chinese. Study III investigates intensity (maximum intensity), another cue to prosodic prominence in Mandarin Chinese. Study IV seeks to further disassociate the relationship between word duration and the maximum pitch/intensity by conducting a series of mediation analyses, including word duration as a fixed effect in Study II and Study III and including maximum pitch/intensity as a fixed effect in Study I.

|  | Duration | Max. Intensity | Intensity Range | Max. Pitch | Pitch Range |
|---|---|---|---|---|---|
| Duration | - | 0.23 | 0.56 | 0.15 | 0.50 |
| Max. Intensity | - | - | 0.28 | 0.12 | 0.18 |
| Intensity Range | - | - | - | 0.17 | 0.31 |
| Max. Pitch | - | - | - | - | 0.64 |
| Pitch Range | - | - | - | - | - |

Table 2: *Correlation matrix of the five acoustic variables showing the Pearson correlation coefficient (R) between the transformed variables.*

Before examining the effect of predictability on acoustic prominence using mixed effects modeling, we also conducted a correlation analysis between the five acoustic variables (after transformation) and the five predictability variables. Results are shown in Table 3. All predictability variables are correlated with all acoustic prominence variables in the expected directions. Frequency, forward predictability and backward predictability are negatively correlated with acoustic prominence. Forward informativity and backward informativity are correlated positively with acoustic prominence.

The raw correlations in Table 3 are useful for diagnosing suppressor effects in our mixed effects models. Suppressor effects, whereby collinearity between two predictors causes the reduction or sign reversal in one of the model estimates, are to be expected, since they were found in Seyfarth (2014) with models containing similarly correlated variables. One diagnostic of a suppressor effect is whether the model estimate is in the same or opposite direction as the correlation between the dependent and independent variable. Model estimates in the opposite direction of the correlation suggest a suppressor effect. We will therefore return to the correlations shown in Table 3 when interpreting the coefficients in our models.

|  | Frequency | Forward predictability (log2) | Backward predictability (log2) | Forward informativity (log2) | Backward informativity (log2) |
|---|---|---|---|---|---|
| Duration | -0.56 | -0.43 | -0.53 | 0.48 | 0.44 |
| Max. Intensity | -0.18 | -0.14 | -0.16 | 0.17 | 0.20 |
| Intensity Range | -0.37 | -0.37 | -0.29 | 0.33 | 0.32 |
| Max. Pitch | -0.06 | -0.05 | -0.05 | 0.04 | 0.06 |
| Pitch Range | -0.29 | -0.21 | -0.27 | 0.25 | 0.23 |

Table 3: Correlation matrix of the acoustic variables by the predictability variables showing the Pearson correlation (R) between the transformed variables.

*3.1 Study 1 -- Duration*

The final fixed and random effects estimates for the duration model are summarized in table 4 and table 5, respectively. All predictors were statistically significant. We first focus on the informativity predictors. The positive estimate for forward informativity indicates that the higher the forward informativity of a word, the longer the word duration (forward informativity: $\beta = 0.0113$, t = 11.65, p < 0.0001). Backward informativity, while significant, shows a suppressor effect since the coefficient is negative (backward informativity: $\beta = -0.0029$, t = -3.73, p < 0.0001) even though the raw correlation is positive (table 3, R = 0.44).

We now examine the control variables. Most of the variables were associated with word duration in the expected direction. Shorter word durations were associated with higher forward and backward predictability, faster speech rate (preceding and following), and words that were previously mentioned (self-mention and cross-speaker mention). Longer word durations were associated with longer word length, neighboring disfluencies (preceding and following) and following pause duration. Male speakers produced shorter word durations. Syntactic category was associated positively with word duration which is unsurprising because it was target coded which uses the mean value of the dependent variable as a predicting value for each category (see Section 2.3.3: Predictor variables).

Two variables were associated with word duration in an unexpected direction: preceding pause duration and word frequency. Unlike English, preceding pause duration was negatively associated with word duration ($\beta = -0.0015$, t= -7.55, p < 0.0001), suggesting a phrasal-initial shortening effect. Phrasal-initial shortening has been previously reported for Mandarin in broadcast news speech data (Liberman, 2014; Yuan, Ryant, & Liberman, 2014) without adjusting for other factors such as word type and tone content. This suggests that our effect is unlikely to be a statistical accident, since it is robust with or without adjusting for other factors and across two speech genres. However, in experimental studies, phrase-initial shortening is less consistent (Tseng, Pin, Lee, Wang, & Chen, 2005; Xu & Wang, 2009; Yang 2011; Yang 2016). Word frequency, while significant, shows a suppressor effect since the coefficient is positive (β = 0.0063, t = 5.96, p < 0.0001) even though the raw correlation is negative (table 4, R = -0.56).

|  | β | SE | t | p |
|---|---|---|---|---|
| Intercept | 2.3443 | 0.0017 | 1366.64 | < 0.0001 |
| Duration baseline | 0.0953 | 0.0010 | 90.12 | < 0.0001 |
| Frequency | 0.0063 | 0.0011 | 5.96 | < 0.0001 |
| Forward predictability | -0.0163 | 0.0003 | -54.72 | < 0.0001 |
| Backward predictability | -0.0189 | 0.0003 | -60.11 | < 0.0001 |
| Forward informativity | 0.0113 | 0.0010 | 11.65 | < 0.0001 |
| Backward informativity | -0.0029 | 0.0008 | -3.73 | 0.0002 |
| Word length | 0.0286 | 0.0009 | 29.96 | < 0.0001 |
| Preceding disfluency = Yes | 0.0558 | 0.0017 | 33.32 | < 0.0001 |
| Following disfluency = Yes | 0.0681 | 0.0025 | 27.52 | < 0.0001 |
| Preceding pause duration | -0.0015 | 0.0002 | -7.55 | < 0.0001 |
| Following pause duration | 0.0432 | 0.0002 | 217.90 | < 0.0001 |
| Preceding speech rate | -0.0105 | 0.0002 | -52.67 | < 0.0001 |
| Following speech rate | -0.0100 | 0.0002 | -50.71 | < 0.0001 |
| Previous self-mention = True | -0.0110 | 0.0005 | -23.72 | < 0.0001 |
| Previous cross-speaker mention = True | -0.0031 | 0.0005 | -6.69 | < 0.0001 |
| Age | 0.0019 | 0.0005 | 3.83 | 0.0001 |
| Gender = Male | -0.0124 | 0.0010 | -12.15 | < 0.0001 |
| Syntactic category | 0.0085 | 0.0006 | 14.59 | < 0.0001 |
| Number of observations: 407,925; number of speakers: 1655; number of word types: 6193 | | | | |

Table 4: *Fixed effect summary for the duration model*

| | SD | Correlation | | |
|---|---|---|---|---|
| Word (intercept) | 0.03258 | - | - | - |
| Speaker (intercept) | 0.01903 | - | - | - |
| Frequency | 0.00215 | 1.00 | - | - |
| Speaker (forward informativity) | 0.01014 | 0.04 | 0.04 | - |
| Speaker (backward informativity) | 0.00439 | 0.00 | 0.00 | -0.30 |
| Residual | 0.11508 | - | - | |

Table 5: *Random effect summary for the duration model*

*3.2 Study 2 -- Pitch*

The final fixed and random effects estimates for the maximum pitch model are summarized in table 6 and table 7, respectively. All except four predictors (Frequency, $p = 0.2520$; Following disfluency, $p = 0.4994$; Age, $p = 0.1164$; Syntactic category, $p = 0.0810$) were statistically significant. Backward informativity was associated with maximum pitch in the expected direction ($\beta = 0.0022$, $t = 4.05$, $p$

< 0.0001). Forward informativity shows a suppressor effect since the coefficient is negative (β = -0.0020, t = -2.95, p = 0.0031) even though the raw correlation is positive (table 4, R = 0.04).

We now examine the control variables. All of the variables were associated with maximum pitch in the expected direction. Lower maximum pitch values were associated with higher forward and backward predictability, faster speech rate (preceding and following), and words that were previously mentioned (self-mention and cross-speaker mention). Higher maximum pitch values were associated with longer word length and preceding disfluencies. The two pause duration variables, preceding pause duration with a positive coefficient and following pause duration with a negative coefficient, suggest that maximum pitch is higher phrase-initially and lower phrase-finally (Xu & Wang, 2009). Male speakers produced lower maximum pitch.

|  | β | SE | t | p |
| --- | --- | --- | --- | --- |
| Intercept | 2.382 | 0.0017 | 1367.46 | < 0.0001 |
| Pitch baseline | 0.0754 | 0.0009 | 80.60 | < 0.0001 |
| Frequency | -0.0008 | 0.0007 | -1.15 | 0.2520 |
| Forward predictability | -0.0041 | 0.0002 | -19.52 | < 0.0001 |
| Backward predictability | -0.0008 | 0.0002 | -3.96 | < 0.0001 |
| Forward informativity | -0.0020 | 0.0007 | -2.95 | 0.0031 |
| Backward informativity | 0.0022 | 0.0005 | 4.05 | < 0.0001 |
| Word length | 0.0043 | 0.0005 | 9.47 | < 0.0001 |
| Preceding disfluency = Yes | 0.0058 | 0.0011 | 4.83 | < 0.0001 |
| Following disfluency = Yes | 0.0012 | 0.0018 | 0.68 | 0.4994 |
| Preceding pause duration | 0.0016 | 0.0001 | 11.30 | < 0.0001 |
| Following pause duration | -0.0008 | 0.0001 | -5.76 | < 0.0001 |
| Preceding speech rate | -0.0030 | 0.0001 | -20.64 | < 0.0001 |
| Following speech rate | -0.0009 | 0.0001 | -6.77 | < 0.0001 |
| Previous self-mention = True | -0.0048 | 0.0003 | -14.24 | < 0.0001 |
| Previous cross-speaker mention = True | -0.0013 | 0.0003 | -3.90 | < 0.0001 |
| Age | -0.0012 | 0.0007 | -1.57 | 0.1164 |
| Gender = Male | -0.1070 | 0.0022 | -47.92 | < 0.0001 |
| Syntactic category | 0.0007 | 0.0004 | 1.75 | 0.0810 |
| Number of observations: 401,629; number of speakers: 1655; number of word types: 6189 | | | | |

Table 6: *Fixed effect summary for the maximum pitch model*

| | SD | Correlation | | |
|---|---|---|---|---|
| Word (intercept) | 0.022278 | - | - | - |
| Speaker (intercept) | 0.031214 | - | - | - |
| Frequency | 0.000828 | 1 | - | - |
| Speaker (forward informativity) | 0.006374 | -0.14 | -0.14 | - |
| Speaker (backward informativity) | 0.00534 | 0.25 | 0.25 | -0.58 |
| Residual | 0.082058 | - | - | - |

Table 7: *Random effect summary for the maximum pitch model*

### 3.3 Study 3 -- Intensity

The final fixed and random effects estimates for the maximum intensity model are summarized in table 8 and table 9, respectively. All except two predictors (*Forward informativity*, p = 0.8251; Age,

p = 0.9919) were statistically significant. While forward informativity did not reach significance, backward informativity was associated with maximum intensity in the expected direction ($\beta = 0.1085$, t = 3.59, p < 0.0001). The higher the backward informativity of a word, the higher the maximum intensity.

We now examine the control variables. Lower maximum intensity values were associated with higher frequency, higher forward and backward predictability, faster speech rate (preceding and following), and words that were previously mentioned (self-mention and cross-speaker mention). Higher maximum intensity values were associated with neighboring disfluencies. The two pause duration variables, preceding pause duration with a positive coefficient and following pause duration with a negative coefficient, suggest that maximum intensity is higher phrase-initially and lower phrase-finally. Syntactic category was associated negatively with maximum intensity.

|  | β | SE | t | p |
|---|---|---|---|---|
| Intercept | 71.2018 | 0.0646 | 1101.47 | < 0.0001 |
| Intensity baseline | 6.6835 | 0.0153 | 438.11 | < 0.0001 |
| Frequency | -0.0959 | 0.0415 | -2.31 | 0.0209 |
| Forward predictability | -0.0703 | 0.0110 | -6.38 | < 0.0001 |
| Backward predictability | -0.1928 | 0.0116 | -16.62 | < 0.0001 |
| Forward informativity | -0.0083 | 0.0375 | -0.22 | 0.8251 |
| Backward informativity | 0.1085 | 0.0303 | 3.59 | 0.0003 |
| Word length | -0.1410 | 0.0252 | -5.59 | < 0.0001 |
| Preceding disfluency = Yes | 0.2743 | 0.0615 | 4.46 | < 0.0001 |
| Following disfluency = Yes | 0.8334 | 0.0897 | 9.29 | < 0.0001 |
| Preceding pause duration | 0.0334 | 0.0072 | 4.65 | < 0.0001 |
| Following pause duration | -0.0333 | 0.0072 | -4.63 | < 0.0001 |
| Preceding speech rate | -0.1120 | 0.0073 | -15.30 | < 0.0001 |
| Following speech rate | -0.1241 | 0.0073 | -17.07 | < 0.0001 |
| Previous self-mention = True | -0.5092 | 0.0171 | -29.68 | < 0.0001 |
| Previous cross-speaker mention = True | -0.2488 | 0.0173 | -14.41 | < 0.0001 |
| Age | -0.0001 | 0.0143 | -0.01 | 0.9919 |
| Gender = Male | 0.1743 | 0.0292 | 5.97 | < 0.0001 |
| Syntactic category | -0.0543 | 0.0238 | -2.28 | 0.0223 |
| Number of observations: 409,735; number of speakers: 1655; number of word types: 6193 | | | | |

Table 8: *Fixed effect summary for the maximum intensity model*

| | SD | Correlation | | | |
|---|---|---|---|---|---|
| Word (intercept) | 1.3339 | - | - | - | - |
| Speaker (intercept) | 0.5117 | - | | | - |
| Frequency | 0.249 | 0.35 | - | - | |
| Speaker (forward informativity) | 0.3069 | 0.19 | 0.32 | - | |
| Speaker (backward informativity) | 0.2644 | -0.07 | 0.3 | -0.01 | |
| Residual | 4.2659 | - | - | - | |

Table 9: *Random effect summary for the maximum intensity model*

*3.4 Study 4 -- Mediation analysis*

We have shown that there is an effect of informativity on duration, pitch and intensity in the last three studies. However, it is possible that these apparent effects could be explained by only one of these dependent variables, specifically the informativity effects on duration, as it is most widely reported, could explain the effects on pitch and intensity.

To evaluate whether the effects of informativity on pitch and intensity could be explained by duration, we conducted mediation analyses by adding duration as a fixed effect to the pitch and intensity models. For completeness, we also fitted the duration model with pitch or intensity as a fixed effect. These mediation analyses are summarized in Table 10. The analyses showed that the two informativity effects remain the same in the three acoustic dimensions even when mediated.

|  |  | Duration | | | Maximum Pitch | | Maximum Intensity | |
|---|---|---|---|---|---|---|---|---|
|  | Mediator | None | Max. Pitch | Max. Intensity | None | Duration | None | Duration |
| Forward informativity | β | 0.0113 | 0.0112 | 0.0122 | -0.0020 | -0.0035 | -0.0083 | -0.1458 |
|  | SE | 0.0010 | 0.0009 | 0.0010 | 0.0007 | 0.0007 | 0.0375 | 0.0368 |
|  | t | 11.65 | 12.03 | 12.62 | -2.95 | -5.19 | -0.22 | -3.96 |
| Backward informativity | β | -0.0029 | -0.0031 | -0.0038 | 0.0022 | 0.0027 | 0.1085 | 0.1419 |
|  | SE | 0.0008 | 0.0008 | 0.0008 | 0.0005 | 0.0005 | 0.0303 | 0.0296 |
|  | t | -3.73 | -4.08 | -4.73 | 4.05 | 5.08 | 3.59 | 4.80 |

Table 10: *Summary of Informativity effects on duration, maximum pitch and maximum intensity*

*3.5 Summary*

Table 11 provides a summary of the main informativity results. Informativity had significant independent effects on three phonetic dimensions related to prosody: pitch, intensity and duration. We focused on max pitch and max intensity because pitch range and intensity range showed stronger correlations with duration; however models fit to pitch range and pitch intensity showed similar patterns with respect to predictability-based variables. These results come from Mandarin Chinese, a language with lexical tone and a well-studied prosodic system. We further showed that the influence of informativity on pitch and intensity was not mediated by duration. We elaborate on the importance of the mediation analysis below. Taken together, the set of informativity results is consistent with the hypothesis that words in the mental lexicon absorb the phonetic reflexes of prosody, reflecting the level of prosodic prominence with which they are typically produced. In short, prosody leaks into the lexicon.

|  | Duration | | | Maximum Pitch | | | Maximum Intensity | | |
| --- | --- | --- | --- | --- | --- | --- | --- | --- | --- |
|  | β | SE | t | β | SE | t | β | SE | t |
| Forward informativity | 0.113 | 0.0010 | 11.65 | -0.0020 | 0.0007 | -2.95 | -0.0083 | 0.0375 | -0.22 |
| Backward informativity | -0.0029 | 0.0008 | -3.73 | 0.0022 | 0.0005 | 4.05 | 0.1085 | 0.0303 | 3.59 |

Table 11: *Informativity summary of duration, maximum pitch and maximum intensity*

## 4 Discussion

Our main result is that informativity, operationalized as the average bigram predictability of a word, influences three phonetic dimensions associated with prosodic prominence in Mandarin Chinese: word duration, maximum pitch and maximum intensity.

Our choice to investigate Mandarin was motivated by several considerations. Importantly, the phonetic characteristics of prosodic prominence are well-known in this language. Another consideration was the general underrepresentation of non-Indo-European languages in research on probabilistic reduction at the phonetic level. The majority of existing research on probabilistic reduction comes from English (e.g., Bell, Brenier, Gregory, Girand, & Jurafsky, 2009), Dutch (Kuperman, Pluymaekers, Ernestus, & Baayen, 2007; Pluymaekers, Ernestus, & Harald Baayen, 2005a, 2005b), French (Bürki, Ernestus, Gendrot, Fougeron, & Frauenfelder, 2011; Pellegrino, Coupé, & Marsico, 2011; Torreira & Ernestus, 2011), Italian (Pellegrino et al., 2011) and Spanish (Torreira & Ernestus, 2012), while there is only a small number of comparable studies on non-Indo-European languages, such as Cantonese (Zhao & Jurafsky, 2009), Japanese (Sano, 2018; Shaw & Kawahara, 2019), and Kaqchikel (Tang & Bennett, 2018). Our results on Mandarin Chinese make a typological contribution to the investigation of probabilistic reduction in the Sino-Tibetan language family. Notably, they come from a language that has lexical tone. Even though pitch is lexically contrastive in Mandarin, we still found significant variation in pitch dictated by predictability and also by informativity.

We interpret these results as resulting from lexicalization of phonetic details associated with prosodic prominence. In the remainder of this discussion, we motivate the close connection that we have assumed between predictability and prosodic prominence (4.1). We then explain how the effect of informativity, a word-level variable, relates to lexicalization (4.2). Lastly, we discuss the relative strength of forward informativity and backwards informativity in the context of cognitive mechanisms proposed to account for the relation between predictability and word duration more broadly (4.3).

*4.1 The connection between predictability and prosodic prominence*

A key assumption behind our interpretation of the results is that predictability conditions prosodic prominence. Here we draw on connections between a few lines of research. The first is prosodic focus, which refers to the emphasis given to words often because they introduce new information to the discourse (Bolinger, 1958) or alternatives to the truth propositional content of an utterance (Rooth, 1992). In Mandarin Chinese (as well as English and many other languages), any number of content words in an utterance can receive prosodic focus, marked phonetically by local increases in maximum pitch, pitch range, duration, intensity, and post-focal compression of intensity and pitch range (e.g., Breen et al. 2010; Cao 2012; Chao 1968; Chen 2006; Chen and Gussenhoven 2008; Chen et al. 2014; Cooper et al. 1985; Ito and Speer 2006; Robert Ladd 2008; Lee et al. 2015; Lieberman 1960; Liu and Xu 2005; Ouyang and Kaiser 2015; Pierrehumbert and Beckman 1988; Xu 1999; Xu and Xu 2005; Xu et al. 2012; Wang and Xu 2011; Yuan 2004). Focused items tend also to be less predictable, as contextual unpredictability is a natural consequence of presenting new information or an item from a set of alternatives. There is a natural link between low predictability and prosodic prominence in focused words. All else equal, new information in a discourse context is likely to receive increased prosodic prominence, as measured in the phonetic signal (e.g., Calhoun, 2010). It is also the case that listeners judge less predictable words to be more prominent, even when bottom-up factors in the speech signal have been controlled for (Cole, Mo, & Hasegawa-Johnson 2010; Bishop, Kuo and Kim (to appear)). In studies of prosodic prominence, both word frequency (Baumann 2014; Cole et al. 2010; Cole et al. 2017; Nenkova et al. 2007) and informativity (Anttila et al., 2018) have been shown to influence prominence perception. This indicates that the connection between predictability and prominence may constitute part and parcel of tacit linguistic knowledge. Taken together, these findings serve to motivate our investigation of predictability, as measured in our study, on phonetic dimensions associated with prosodic prominence.

The link between predictability and prosodic prominence is closely related to the Uniform Information Density (UID) hypothesis (Jaeger, 2010) and the Smooth Signal Redundancy (SSR) hypothesis (Turk, 2010). We discuss each of these hypotheses in turn.

The UID proposes that predictability influences selection between grammatically available variants at all levels of linguistic structure. When a choice is made available by the grammar, speakers choose variants that maintain uniform distribution of information across some dimension (time, linear order of words, effort, etc.). Prosody, as a level of linguistic structure, would presumably fall within the scope of the UID as proposed by Jaeger (2010). Our hypothesis could therefore be viewed as a specific application of UID to prosodic structure; that is, prosodic structure, including phrasing and levels of prominence, are selected to distribute information uniformly across time. This proposal refers to a specific level of linguistic structure, prosody, and a specific dimension, time, over which information is distributed uniformly. The general version of UID does not commit to time as the dimension over which information is uniformly distributed. For example, the UID is consistent as well with production effort as the dimension that trades off with information. Speech could be varied so that information is distributed across speaker effort, i.e., less production effort invested in less informative words, syllables, segments. Xu & Prom-on (2019) provide a critical evaluation of the "effort" hypothesis, arguing that, time, not energy, is the most valuable resource for speech production; on their proposal, the amount of time allocated to each linguistic unit is a function of its importance. This predicts the relation between predictability and duration found in our study, but it does not necessarily predict effects of predictability on pitch and intensity, unless of course they are mediated by variation in duration. As words get shorter, the time required to achieve a pitch target could be reduced to the point that that the target is not achieved, i.e., target undershoot (Lindblom 1963). Crucially, our mediation analysis in section 4.3 shows that the influences that informativity has on pitch and intensity are not mediated by duration.

The Smooth Signal Redundancy (SSR) hypothesis states that prosody modulates the speech signal to maintain stable word recognition probability over time (Aylett & Turk, 2004; Turk, 2010). Part of recognition probability comes from predictability, or language redundancy; part comes from the speech signal. The connection between predictability and prosodic prominence discussed above is highly consistent with SSR. SSR dictates that predictability (as an index of language redundancy) trades off with signal enhancement, which includes prosodic prominence. Thus, the key prediction of our hypothesis is one that is shared by SSR—phonetic dimensions under prosodic control are precisely those that are influenced by contextual probability.

To summarize, the preceding discussion motivates a connection between prosody and predictability. Since prosodic phrasing and prominence affects not only the phonetic dimension of word duration but also pitch and intensity, we explored whether effects of predictability found on duration in past studies would generalize beyond word duration to pitch and intensity, a prediction that was confirmed in our data.

*4.2 Reflections of prosodic prominence in the lexicon*

In addition to the effects of predictability, we also found significant influences of the average predictability of a word –, i.e., informativity – on word duration, pitch and intensity. It is on the basis of this result that we make our second claim, that prosodic influences on words are absorbed into the lexicon. Informativity, as it has been computed in this study and elsewhere, is a word-level variable. Each word has a unique set of contexts in which it occurs. Within these contexts, a word can be more or less predictable. Our results show that each word also has a characteristic prosodic profile, represented by combinations of duration, pitch, and intensity. Crucially, a word's prosodic profile is systematically related to the contexts in which it occurs. This was captured in our models through the effect of informativity on word duration, pitch, and intensity.

The most general version of our claim is that words come to take on the phonetic characteristics of the prosodic contexts in which they are typically produced. Words that typically occur in prosodically prominent positions are correspondingly produced with greater duration, pitch and intensity. We view this second claim as compatible with a range of perspectives on the lexicon. Minimally, it requires that phonetic representations associated with words can change over time, a fact that is well-established (e.g., Hay et al., 2015, Sóskuthy & Hay 2017). Theories that adopt a lexicon of phonetically-detailed episodic memories capture changes in the lexicon over time rather directly, through combinations of memory trace decay and the addition of new exemplars (Goldinger, 1998; e.g., Johnson, 1997; Pierrehumbert, 2001; Foulkes & Docherty, 2006). In these theories, change in the speech community over time is directly encoded in the lexicon, at least to the extent that the dimensions of variation are encoded veridically (Foulkes & Docherty 2006; Foulkes & Hay, 2015). Accordingly, the prosodic prominence with which a word is produced will come to shape the long term representation of that word, as shown in Sóskuthy & Hay (2017), alongside other aspects of a word's context, including the typical location that it was produced (Hay et al. 2017) and the speech characteristics of typical users of a word (Walker & Hay 2011).

Lexical representations that are somewhat more abstract in that they disentangle various influences on the speech signal can also be seen to make the same prediction. This is because listener attribution of the source of phonetic variation is often imperfect. Consider for example a word produced in a low predictability context with a correspondingly high degree of phonetic prominence. A listener may attribute some degree of prominence to the particular context in which the word was produced, and represent only residual phonetic details as associated with the lexical item. This process involves some degree of abstraction in that the details of the word's pronunciation are abstracted away from the particular context. However, on this account, prosodic prominence can still influence the long-term representation of words. This is because listener compensation for contextual effects is typically imperfect. A well-studied case is compensation for coarticulation. Listeners routinely attribute some aspects of the phonetic signal associated with a speech segment to the coarticulatory context in which it was produced (e.g., Beddor et al., 2003); however, such compensation is typically incomplete (Cole et al., 2011) and varies in degree across listeners (Yu et al., 2015), a set of facts which has also received a rational analysis (Sonderegger and Yu, 2010). Incomplete compensation for the influence of prosodic prominence on word forms makes the same prediction as "pure" episodic representation of words. Over time, the lexicon will come to reflect the prosodic ecology of language use. This basic idea is broadly compatible with different conceptualizations of the mental lexicon.

Words that are typically produced in prominent environments will come to take on the phonetic characteristics of prominent environments, even when produced in prosodically weak positions. To the extent that predictability drives prosodic prominence, an assumption we motivated in 4.1, this prediction is borne out in the data as a significant effect of informativity on the phonetic dimensions of prosodic prominence in Mandarin Chinese: pitch, duration and intensity.

There is already some evidence for the lexical encoding of prosodic patterns coming from studies on German and English, languages in which pitch patterns (or tunes) are assigned at the phrasal level (Calhoun & Schweitzer, 2012; Schweitzer et al., 2015). Schweitzer et al. showed that f0 contours are more stable in predictable collocations than in unpredictable collocations, suggesting a possible lexicalization of intonation. Sóskuthy & Hay (2017) showed that words that tend to occur at the ends of intonational phrases are longer, even when they occur in other environments, another case of prosodically conditioned phonetic variation leaking into the lexicon. In a study predicting the presence/absence of a phrasal pitch accent in a manually labelled portion of the Switchboard corpus, Nenkova et al. (2007) showed that a lexical property, the probability of particular lexical items to bear accent, was by far the best predictor (explaining 75.59% of the data). Anttila et al. (to appear) examined a finer-grained annotation of English prominence. Their corpus was hand-annotated for metrical grids encoding gradient levels of sentence prominence. Results revealed that nouns had much higher levels of prominence than verbs and function words and that informativity was a significant predictor of prominence judgments even after grammatical factors, e.g., Nuclear Stress Rule (Chomsky & Halle, 1968), had been taken into account. These results are consistent with the view that sentence-level prominence may be a driving force in shaping word-level stress (Anttila, Dozat, Galbraith, & Shapiro, to appear).

In sum, prosody leaves its imprint on the phonetic form of words, and the mental lexicon reflects the typical prosodic contexts in which words are produced.

*4.3 The broader context of our proposal and relation to other accounts*

One empirical contribution of the current paper is that we show that predictability and informativity influence pitch and intensity, in addition to duration, and that the effects of informativity on pitch and intensity are not mediated by duration. This is theoretically significant in part because it also narrows the range of possible accounts to eliminate those that predict only effects of information on duration.

There is a substantial body of work exposing systematic relations between "speech rate", as quantified by the duration of linguistic units, e.g., segments, syllables, words, and the information contained in those units (Arnon & Cohen Priva, 2015; Arnon & Priva, 2013; Arnon & Snider, 2010; Aylett & Turk, 2004, 2006; Bell et al., 2009; Jurafsky et al., 2001; Priva, 2015; Shaw & Kawahara, 2019; Tang & Bennett, 2018). This work points to a tradeoff between time and information – linguistic units carrying more information tend to take more time to produce. Another dimension of variation is vowel and syllable reduction (e.g., Jurafsky et al. 2001; Aylett & Turk, 2004), although it is also the case that shorter vowel durations can condition vowel reduction—as the movement of speech organs may fail to achieve their targets under time pressure, i.e., "target undershoot" (Lindlom 1963; Lindblom & Moon, 1993). Thus, the empirical basis of much of the existing theorizing about predictability in speech is based on duration or potentially duration-mediated factors, such as vowel reduction. Notable exceptions include Watson et al. (2008), who evaluate effects of predictability and context importance on duration, intensity and pitch in an experimental game-like setting and Fitzroy and Breen (2020) who compare effects of predictability on intensity with previous results on duration (Breen 2020). Both of these studies reveal differential effects of predictability on duration and other dimensions of prosodic prominence.

Focusing just on duration patterns, relatively consistent information rates—achieved by trading off time-per-unit with information-per-unit—have been observed across languages (Coupé, Oh, Dediu, & Pellegrino, 2019), across speakers of the same language (Cohen Priva, 2017), and within speakers in different situations (Arnold, Bennetto, & Diehl, 2009; Buz, Tanenhaus, & Florian Jaeger, 2016; Kitamura, 2014; Maniwa, Jongman, & Wade, 2008; Raveh, Steiner, Siegert, Gessinger, & Möbius, 2019; Schertz, 2013). The ubiquity of this pattern across these levels of description suggests a cognitive basis for the behavior. That is, the production mechanism at the level of the individual constrains speech to adhere to relatively stable information rates. Consequently, the same patterns found within the individual can be found on average within a speech community and across speech communities.

To the extent that these patterns are ubiquitous across languages, including observance at the level of individual speakers, an account rooted in fundamental aspects of human language, as embodied in the mind/brain, is justified. "Efficient communication" is often evoked in this context as a universal functional constraint on language. Languages, sitting at the intersection of biology, ecology and culture, have in common that they evolve to serve a communicative function (Gibson et al., 2019; Winter & Christiansen, 2012). However, functional pressures on the development of the system are distinct from the internal workings. Cognitive mechanisms that have been proposed to explain stable information rates include those that are largely situated within the speech production system proper (Bell et al., 2009), and those that evoke "audience design", a language-specific application of theory of mind (Arnold et al., 2009; Arnold, Kahn, & Pancani, 2012; Watson et al., 2008, 2010). Our proposal is that the assignment of prosodic prominence is the primary driver of the patterns.

In assessing the degree to which our account is compatible with those put forward to explain duration-based patterns, it is useful to compare the results on forward vs. backward predictability/informativity. Forward predictability indicates how predictable a word is from its preceding context. A word with high forward predictability is predictable from what comes before it. This type of predictability is useful in speech perception, on the common assumption that listeners actively narrow the field of competitors based on preceding context (Marslen-Wilson & Welsh, 1978; McClelland & Elman, 1986; Norris, 1994). Listeners also make backward inferences, using information that comes later in time to resolve earlier uncertainty (e.g., Toscano & McMurray 2010), and there is evidence that both backward and forward predictability, defined at the level of bigrams, are relevant to Chinese word formation processes, with forward predictability having a stronger effect than backward predictability (Shaw et al., 2014). Backward inference in perception is generally slower and less efficient than forward inference (Nooteboom, 1981). Varying word forms according to listener needs (audience design) would thus entail phonetic variation, i.e., enhancement/reduction, conditioned by forward predictability. Backward predictability, on the other hand, is consistent with production-based accounts of probabilistic reduction. A speaker typically plans chunks of speech consisting of multiple words. Consider a sequence of words, AB, in the production plan. If A is predictable from B, then A can be retrieved from the lexicon more easily. The speaker has access to both A and B in speech planning while the listener receives information more linearly, having to wait to hear A before getting information about B (modulo any effects of anticipatory coarticulation). For this reason, backward informativity has been more closely linked to production-based accounts of probabilistic reduction.

An example of a production-based account is that lexical items that are harder to retrieve are also produced more slowly (Bell et al., 2009). Coupling the time course of lexical access with the speed of word production is possibly crucial to fluent speech. Hesitations or pauses would result if lexical access lags behind speech rate; lexical access outpacing speech rate could lead to pathological coarticulation or anticipatory substitution errors, both of which are well-documented speech anomalies (Cutler, 1982; Dell, 1986; Frisch & Wright, 2002; Fromkin, 1984; Goldstein, Pouplier, Chen, Saltzman, & Byrd, 2007; McMillan & Corley, 2010). Considering that a larger sequence of words is available for production planning, it is reasonable to assume that backward predictability may aid lexical access in speech production. In the other studies to date reporting effects of informativity on word duration, it was backward informativity that showed reliable effects across corpora (Seyfarth, 2014; Sóskuthy & Hay, 2017). This may suggest that, in English, it is the residue of lexical access in speech production that is lexicalized as word-specific phonetic durations.

In our study, we found significant negative effects—the expected direction—of both forward and backward predictability on all three phonetic variables: duration, pitch and intensity. In light of this result and the established connections between predictability and prosodic prominence (Section 4.1), we speculate that the assignment of prosodic prominence in Mandarin Chinese is sensitive to both forward and backward predictability. However, we cannot completely rule out the possibility that some aspects of lexical access difficulty—per the production account—may have also left their imprint on our spontaneous speech data.

A number of factors are known to influence both lexical access in speech production and word duration, including word frequency, phonological neighborhood density (Gahl & Strand, 2016; Gahl, Yao, & Johnson, 2012) and metrical predictability (Breen, 2018; Shaw, 2013). The dual effects of such factors on both lexical access and speech rate are consistent with the proposal that speech rate is yoked to lexical access or "production ease". Whether or not there is a causal relation between lexical access and word duration or whether these factors operate on lexical access and word production independently is an area that requires future research. Work on one of these factors, phonological neighborhood density, has found that its influence on speech rate (word duration) is independent of its influence on lexical access, which challenges some versions of the production ease account (Buz & Jaeger, 2016). This still leaves open the possibility that backward predictability facilitates lexical access in spontaneous speech production and that this has a knock-on effect on word duration, which may be independent of prosodic prominence. Other studies as well have concluded that a complete account of word prominence variation likely involves multiple cognitive factors (e.g., Lam & Watson 2010; Fitzroy & Breen 2019). It is possible as well that in our own data the significant effect of backward predictability on duration is indexing some aspects of lexical access difficulty. However, this would not explain why backward predictability also influences pitch and intensity.

When it comes to the lexicalization of local influences on pitch, intensity, and duration, we found that not all predictability effects surface also as informativity effects. That is, the lexicalization of local influences on prosodic prominence is selective. As we argued in section 4.2, informativity provides a test of lexicalization. Even though we observed local effects of both forward and backward predictability on all three phonetic dimensions, only some of these predictability effects surfaced as informativity effects. For word duration, there was a significant positive effect—the expected direction—only for forward informativity. The absence of this effect for backward informativity indicates that local influences of backward predictability on word duration does not get lexicalized, at least not in Mandarin Chinese. For pitch and intensity, the directionality was different—there was a significant positive effect only for backward informativity and not forward informativity. Thus, while the data support the broad claim that words in the lexicon adapt to their usage context, and, more specifically, to their prosodic context, some effects are more likely than others to work their way into the lexicon.

Conceivably, the pattern of informativity effects reveals something about how context influences listener attention to phonetic detail. That is, in Mandarin Chinese, the preceding context (forward predictability) may condition listener attention to word duration while the following context (backward predictability) may condition attention to pitch and intensity. For the following context to actively contribute to word recognition, the phonetic details of a target word would have to be held in short-term memory long enough for the following word to be recognized. Increased pitch/intensity may direct attention to phonetic detail and facilitate retention long enough to benefit from the following context. Increased word duration, in contrast, delays the onset of the following word. From this standpoint, it is reasonable that increased word duration better facilitates recognition conditioned on the preceding context (forward informativity), because it provides more time to integrate phonetic details of the target word with a known context. Such effects fall under the broader hypothesis that the likelihood of lexicalization of some phonetic dimension is related to the role that dimension plays in spoken word recognition. Phonetic dimensions that facilitate word recognition may be more likely to be lexicalized. We are not aware of any direct tests of this hypothesis but we see it as a fruitful direction for future research.

In summary, our results bear directly on recent debates about the cognitive mechanism(s) responsible for the observation that word length/duration correlates with contextual predictability; see Jaeger and Buz (2017) for an overview. One perspective is that speakers actively balance contextual predictability with signal redundancy (e.g., Jaeger, 2015; Turk 2010; Wedel, Nelson, & Sharp, 2018) possibly driven by audience design (Watson, Arnold, & Tanenhaus, 2008, 2010). Fewer resources in production are expended when communication is not at risk (predictable contexts); additional production resources are drawn upon in challenging communication environments (unpredictable contexts, noisy environments, etc.). A key characteristic of audience design accounts is that speakers adjust pronunciation based on an internal model of listener perception (Rosa, Finch, Bergeson, & Arnold, 2015). The internal model informs speakers of how words are likely to be understood given the context. Speakers use this knowledge to modulate the phonetic signal to facilitate listener recognition of the intended message.

One important aspect of our result is that effects of predictability were found not just on word duration but also on other dimensions of prosodic prominence that are not directly mediated by word duration. We have argued that this supports a prosodic account. Speakers vary prosodic prominence according to the local (forward and backward) predictability of words. A second important aspect of the result is that a word-based variable, informativity, predicted variation in duration, pitch, and intensity above and beyond the effects of local predictability. These informativity effects were not uniform across preceding and following contexts. Rather, positive informativity effects on word duration were based on the preceding context (forward informativity) while informativity effects on pitch and intensity were based on the following context (backward informativity). We have argued on the basis of this result that prosodic influences on words affect long-term memory, reflected in speech production. We view this account in terms of prosodic prominence as consistent with audience design. Future research may show that the assignment of prosodic prominence is tied to speaker rendering of interlocutor mental state. In this case, the assignment of prosodic prominence may provide a mechanism through which audience design operates. The strongest version of this hypothesis makes the prediction that the resources available to audience design considerations are constrained by the resources of prosody in language-specific ways. If so, deeper evaluation of these cognitive mechanisms must be pursued in the context of the linguistic systems in which they operate.

## 5 Conclusion

Phonetic correlates of prosodic prominence in Mandarin Chinese, pitch, intensity, and duration, were shown to vary with the average predictability of a word in context, i.e., the word's informativity. The sensitivity of phonetic dimensions associated with prosody to predictability underscores a relation identified in past work—less predictable words tend to attract prosodic prominence. More importantly, the influence of a word's informativity on the phonetic dimensions of prosody indicates that the level of prominence with which a word is typically produced may influence its lexical representation. That is, the long-term representation of a lexical item takes on the phonetic characteristics of the prosodic context in which it typically occurs. This result builds on a substantial body of work establishing phonetic cues to prosodic structure in Mandarin Chinese, a language with lexical tone. More broadly, the findings suggest that the phonetics of a prosodic system can contribute as well to an understanding of phonetic variation in the lexicon.

## 6 Acknowledgement

We thank audiences at Annual Meeting on Phonology 2019, New Observations in Speech and Hearing seminar at Ludwig Maximilian University of Munich, Lunchtime talk series at Yale University, Brain and Language seminar at University of Florida, and Wayne State University. We thank Jason Bishop for extensive discussion of prosodic, and Ratree Wayland, Márton Sóskuthy and one anonymous reviewer for comments which greatly improved the paper. We also thank Xue Liang Chen for manually checking the lexical items in the acoustic corpus for alternative pronunciations and Ruoying Zhao for discussing the part of speech tagging system.

## 7 References

Adelman, J. S., Brown, G. D. A., & Quesada, J. F. (2006). Contextual diversity, not word frequency, determines word-naming and lexical decision times. *Psychological Science*, *17*(9), 814–823.

Anttila, A., Dozat, T., Galbraith, D., & Shapiro, N. (To appear). Sentence stress in presidential speeches. In G. Kentner & J. Kremers (Eds.), *Prosody in Syntactic Encoding, special issue of Linguistische Arbeiten*.

Arnold, J. E. (2008). THE BACON not the bacon: How children and adults understand accented and unaccented noun phrases. *Cognition*, Vol. 108, pp. 69–99. https://doi.org/10.1016/j.cognition.2008.01.001

Arnold, J. E., Bennetto, L., & Diehl, J. J. (2009). Reference production in young speakers with and without autism: effects of discourse status and processing constraints. *Cognition*, *110*(2), 131–146.

Arnold, J. E., Kahn, J. M., & Pancani, G. C. (2012). Audience design affects acoustic reduction via production facilitation. *Psychonomic Bulletin & Review*, *19*(3), 505–512.

Arnon, I., & Cohen Priva, U. (2015). Time and again: The changing effect of word and multiword frequency on phonetic duration for highly frequent sequences. *The Mental Lexicon*, *9*(3), 377–400.

Arnon, I., & Priva, U. C. (2013). More than words: the effect of multi-word frequency and constituency on phonetic duration. *Language and Speech*, *56*(Pt 3), 349–371.

Arnon, I., & Snider, N. (2010). More than words: Frequency effects for multi-word phrases. *Journal of Memory and Language*, Vol. 62, pp. 67–82. https://doi.org/10.1016/j.jml.2009.09.005

Assaneo, M. F., Florencia Assaneo, M., & Poeppel, D. (2018). The coupling between auditory and motor cortices is rate-restricted: Evidence for an intrinsic speech-motor rhythm. *Science Advances*, Vol. 4, p. eaao3842. https://doi.org/10.1126/sciadv.aao3842

Aylett, M., & Turk, A. (2004). The smooth signal redundancy hypothesis: a functional explanation for relationships between redundancy, prosodic prominence, and duration in spontaneous speech. *Language and Speech*, *47*(Pt 1), 31–56.

Aylett, M., & Turk, A. (2006). Language redundancy predicts syllabic duration and the spectral characteristics of vocalic syllable nuclei. *The Journal of the Acoustical Society of America*, *119*(5 Pt 1), 3048–3058.

Barr, D. J., Levy, R., Scheepers, C., & Tily, H. J. (2013). Random effects structure for confirmatory hypothesis testing: Keep it maximal. *Journal of Memory and Language*, *68*(3). https://doi.org/10.1016/j.jml.2012.11.001

Bates, D., Alday, P., Kleinschmidt, D., Calderón, S., Noack, A., Likan Zhan, Arslan, A., Bouchet-Valat, M., Kelman, T., Baldassari, A., Benedikt Ehinger, Saba, E., Hatherly, M., Morten Piibeleht, Patrick Kofod Mogensen, Babayan, S., & Yakir Luc Gagnon. (2020, October). JuliaStats/MixedModels.jl: v3.0.0. Github. https://github.com/JuliaStats/MixedModels.jl/releases/tag/v3.0.0

Bell, A., Brenier, J. M., Gregory, M., Girand, C., & Jurafsky, D. (2009). Predictability effects on durations of content and function words in conversational English. *Journal of Memory and Language*, Vol. 60, pp. 92–111. https://doi.org/10.1016/j.jml.2008.06.003

Bell, A., Jurafsky, D., Fosler-Lussier, E., Girand, C., Gregory, M., & Gildea, D. (2003). Effects of disfluencies, predictability, and utterance position on word form variation in English conversation. *The Journal of the Acoustical Society of America*, *113*(2), 1001–1024.

Bezanson, J., Edelman, A., Karpinski, S., & Shah, Viral B. (2017). Julia: A fresh approach to numerical computing. SIAM Review, 59, 65–98. https://doi.org/10.1137/141000671

Biber, D. (1993). Representativeness in Corpus Design. *Literary and Linguistic Computing*, *8*(4), 243–257.

Bigi, B. (2015). SPPAS - Multi-lingual Approaches to the Automatic Annotation of Speech. *The Phonetician -- International Society of Phonetic Sciences*, *2015-I-II*(111-112 ), 54–69.

Boersma, P., & Weenink, D. (2019). Praat: doing phonetics by computer [Computer program] (Version 6.0.49). Retrieved from http://www.praat.org/

Bolinger, D. L. (1958). Stress and Information. *American Speech*, Vol. 33, p. 5. https://doi.org/10.2307/453459

Bredart, S. (2002). *The Cognitive Psychology of Proper Names*. https://doi.org/10.4324/9780203132364

Breen, M. (2018). Effects of metric hierarchy and rhyme predictability on word duration in The Cat in the Hat. *Cognition*, *174*, 71–81.

Breen, M., Fedorenko, E., Wagner, M., & Gibson, E. (2010). Acoustic correlates of information structure. *Language and Cognitive Processes*, Vol. 25, pp. 1044–1098. https://doi.org/10.1080/01690965.2010.504378

Brysbaert, M., & Boris New. (2009). Moving beyond Kučera and Francis: A critical evaluation of current word frequency norms and the introduction of a new and improved word frequency measure for American English. *Behavior Research Methods*, Vol. 41, pp. 977–990. https://doi.org/10.3758/brm.41.4.977

Bürki, A., Ernestus, M., & Frauenfelder, U. H. (2010). Is there only one "fenêtre" in the production lexicon? On-line evidence on the nature of phonological representations of pronunciation variants for French schwa words. *Journal of Memory and Language*, Vol. 62, pp. 421–437. https://doi.org/10.1016/j.jml.2010.01.002

Bürki, A., Ernestus, M., Gendrot, C., Fougeron, C., & Frauenfelder, U. H. (2011). What affects the presence versus absence of schwa and its duration: a corpus analysis of French connected speech. *The Journal of the Acoustical Society of America*, *130*(6), 3980–3991.

Buz, E., & Jaeger, T. F. (2016). The (in)dependence of articulation and lexical planning during isolated word production. *Language, Cognition and Neuroscience*, Vol. 31, pp. 404–424. https://doi.org/10.1080/23273798.2015.1105984

Buz, E., Tanenhaus, M. K., & Florian Jaeger, T. (2016). Dynamically adapted context-specific hyper-articulation: Feedback from interlocutors affects speakers' subsequent pronunciations. *Journal of Memory and Language*, Vol. 89, pp. 68–86. https://doi.org/10.1016/j.jml.2015.12.009

Bybee, J. L., & Hopper, P. (2001). *Frequency and the Emergence of Linguistic Structure*. John Benjamins Publishing.

Cai, Q., & Brysbaert, M. (2010). SUBTLEX-CH: Chinese Word and Character Frequencies Based on Film Subtitles. *PLoS ONE*, Vol. 5, p. e10729. https://doi.org/10.1371/journal.pone.0010729

Calhoun, S. (2010). The centrality of metrical structure in signaling information structure: A probabilistic perspective. *Language*, *86*(1), 1–42.

Calhoun, S., & Schweitzer, A. (2012). Can intonation contours be lexicalised? Implications for discourse meanings. *Prosody and Meaning*, 271–328.

Cao, J. (2012). Pitch prominence and tonal typology for low register tone in Mandarin. *Tonal Aspects of Languages-Third International Symposium*. Retrieved from https://www.isca-speech.org/archive/tal_2012/tl12_O3-05.html

Cerda, P., Varoquaux, G., & Kégl, B. (2018). Similarity encoding for learning with dirty categorical variables. *Machine Learning*, Vol. 107, pp. 1477–1494. https://doi.org/10.1007/s10994-018-5724-2

Chao, Y. R. (1968). *A Grammar of Spoken Chinese. Berkeley and Los Angeles: Univ*. of California Press.

Chen, S. F., and Goodman, J. (1999). "An empirical study of smoothing techniques for language modeling," Comput. Speech Lang.13 (4), 359–393.

Chen, Y. (2006). Durational adjustment under corrective focus in Standard Chinese. *Journal of Phonetics*, *34*(2), 176–201.

Chen, Y., & Gussenhoven, C. (2008). Emphasis and tonal implementation in Standard Chinese. *Journal of Phonetics*, *36*(4), 724–746.

Chen, Y., Xu, Y., & Guion-Anderson, S. (2014). Prosodic Realization of Focus in Bilingual Production of Southern Min and Mandarin. *Phonetica*, *71*(4), 249–270.

Chomsky, N., & Halle, M. (1968). *The Sound Pattern of English*. New York: Harper and Row.

Cohen, G., & Burke, D. M. (1993). Memory for proper names: A review. *Memory*, Vol. 1, pp. 249–263. https://doi.org/10.1080/09658219308258237

Cohen Priva, & Jaeger, F. T. (2018). The interdependence of frequency, predictability, and informativity in the segmental domain. *Linguistics Vanguard*, *4*(2). https://doi.org/10.1515/lingvan-2017-0028

Cohen Priva, U. (2015). Informativity affects consonant duration and deletion rates. *Laboratory Phonology*, Vol. 6. https://doi.org/10.1515/lp-2015-0008

Cohen Priva, U. (2017). Not so fast: Fast speech correlates with lower lexical and structural information. *Cognition*, *160*, 27–34.

Cole, J., Mo, Y., & Hasegawa-Johnson, M. (2010a). Signal-based and expectation-based factors in the perception of prosodic prominence. *Laboratory Phonology*, *1*(2), 425–452. https://doi.org/10.1515/labphon.2010.022

Cooper, W. E., Eady, S. J., & Mueller, P. R. (1985). Acoustical aspects of contrastive stress in question–answer contexts. *The Journal of the Acoustical Society of America*, *77*(6), 2142–2156.

Coupé, C., Oh, Y., Dediu, D., & Pellegrino, F. (2019). Different languages, similar encoding efficiency: Comparable information rates across the human communicative niche. *Science Advances*, Vol. 5, p. eaaw2594. https://doi.org/10.1126/sciadv.aaw2594

Cutler, A. (Ed.). (1982). *Slips of the Tongue and Language Production*. Walter de Gruyter.

Daland, R., & Zuraw, K. (2018). Loci and locality of informational effects on phonetic implementation. *Linguistics Vanguard*, *4*(2). Retrieved from doi.org/10.1515/lingvan-2017-0045

de Jong, & Wempe, T. (2009). Praat script to detect syllable nuclei and measure speech rate automatically. *Behavior Research Methods*, *41*(2), 385–390.

Dell, G. S. (1986). A spreading-activation theory of retrieval in sentence production. *Psychological Review*, Vol. 93, pp. 283–321. https://doi.org/10.1037//0033-295x.93.3.283

Denisowski, P. A. (2018). *CC-CEDICT* [Data set]. Retrieved from https://www.mdbg.net/chinese/dictionary?page=cc-cedict

Dilley, L. C., & McAuley, J. D. (2008). Distal prosodic context affects word segmentation and lexical processing. *Journal of Memory and Language*, Vol. 59, pp. 294–311. https://doi.org/10.1016/j.jml.2008.06.006

Drager, K. (2011). Speaker age and vowel perception. Language and Speech, 54(1), 99-121.

Duanmu, S. (2007). *The Phonology of Standard Chinese*. Oxford University Press.

Ellis, D. (2011). *Objective measures of speech quality/SNR*. Retrieved from http://labrosa.ee.columbia.edu/projects/snreval/

Fitzroy, A. B., & Breen, M. (2019). Metric Structure and Rhyme Predictability Modulate Speech Intensity During Child-Directed and Read-Alone Productions of Children's Literature. *Language and Speech*, 0023830919843158.

Foulkes, P., & Hay, J. B. (2015). The Emergence of Sociophonetic Structure. In B. WacWhinney & W. O'Grady (Eds.), The handbook of language emergence (Vol. 87, pp. 292-313). Oxford: Blackwell.

Fowler, C. A. (1988). Differential Shortening of Repeated Content Words Produced in Various Communicative Contexts. *Language and Speech*, Vol. 31, pp. 307–319. https://doi.org/10.1177/002383098803100401

Fowler, C. A., & Housum, J. (1987). Talkers' signaling of "new" and "old" words in speech and listeners' perception and use of the distinction. *Journal of Memory and Language*, Vol. 26, pp. 489–504. https://doi.org/10.1016/0749-596x(87)90136-7

Frisch, S. A., & Wright, R. (2002). The phonetics of phonological speech errors: An acoustic analysis of slips of the tongue. *Journal of Phonetics*, Vol. 30, pp. 139–162. https://doi.org/10.1006/jpho.2002.0176

Fromkin, V. A. (1984). *Speech Errors as Linguistic Evidence*. Walter de Gruyter.

Fung, P., Huang, S., & Graff, D. (2005). *HKUST Mandarin Telephone Speech, Part 1. LDC2005S15. Web Download*.

Gahl, S. (2008). *Time* and *Thyme* Are not Homophones: The Effect of Lemma Frequency on Word Durations in Spontaneous Speech. *Language*, Vol. 84, pp. 474–496. https://doi.org/10.1353/lan.0.0035

Gahl, S., & Strand, J. F. (2016). Many neighborhoods: Phonological and perceptual neighborhood density in lexical production and perception. *Journal of Memory and Language*, Vol. 89, pp. 162–178. https://doi.org/10.1016/j.jml.2015.12.006

Gahl, S., Yao, Y., & Johnson, K. (2012). Why reduce? Phonological neighborhood density and phonetic reduction in spontaneous speech. *Journal of Memory and Language*, Vol. 66, pp. 789–806. https://doi.org/10.1016/j.jml.2011.11.006

Gibson, E., Futrell, R., Piantadosi, S. T., Dautriche, I., Mahowald, K., Bergen, L., & Levy, R. (2019). How Efficiency Shapes Human Language. *Trends in Cognitive Sciences*, *23*(12), 1087.

Goldinger, S. D. (1998). Echoes of echoes? An episodic theory of lexical access. *Psychological Review*, *105*(2), 251–279.

Goldrick, M., Keshet, J., Gustafson, E., Heller, J., & Needle, J. (2016). Automatic analysis of slips of the tongue: Insights into the cognitive architecture of speech production. *Cognition*, Vol. 149, pp. 31–39. https://doi.org/10.1016/j.cognition.2016.01.002

Goldstein, L., Pouplier, M., Chen, L., Saltzman, E., & Byrd, D. (2007). Dynamic action units slip in speech production errors. *Cognition*, *103*(3), 386–412.

Gries, S. T. (2008). Dispersions and adjusted frequencies in corpora. *International Journal of Corpus Linguistics*, Vol. 13, pp. 403–437. https://doi.org/10.1075/ijcl.13.4.02gri

Gries, S. T. (2010). Dispersions and adjusted frequencies in corpora: further explorations. *Corpus-Linguistic Applications*. https://doi.org/10.1163/9789042028012_014

Harrell Jr, F. E. (2015). Regression modeling strategies: with applications to linear models, logistic and ordinal regression, and survival analysis. Springer.

Hall, K. C., Hume, E., Jaeger, T. F., & Wedel, A. (2018). The role of predictability in shaping phonological patterns. *Linguistics Vanguard*, 4(s2).

Hay, J., Podlubny, R., Drager, K., & McAuliffe, M. (2017). Car-talk: Location-specific speech production and perception. *Journal of Phonetics*, *65*, 94-109.

Hsu, B.-J. (2009). MIT Language Modeling Toolkit 0.4.1, http://code.google.com/p/mitlm/.

Howell, P., & Vause, L. (1986). Acoustic analysis and perception of vowels in stuttered speech. *The Journal of the Acoustical Society of America*, Vol. 79, pp. 1571–1579. https://doi.org/10.1121/1.393684

Howell, P., & Williams, M. (1992). Acoustic analysis and perception of vowels in children's and teenagers' stuttered speech. *The Journal of the Acoustical Society of America*, Vol. 91, pp. 1697–1706. https://doi.org/10.1121/1.402449

Ip, M. H. K., & Cutler, A. (2020). Universals of listening: Equivalent prosodic entrainment in tone and non-tone languages. *Cognition*, 202, 104311.

Ito, K., & Speer, S. R. (2006). Using interactive tasks to elicit natural dialogue. *Methods in Empirical Prosody Research*, 229–257.

Ivanova, I., Salmon, D. P., & Gollan, T. H. (2013). The Multilingual Naming Test in Alzheimer's Disease: Clues to the Origin of Naming Impairments. *Journal of the International Neuropsychological Society*, Vol. 19, pp. 272–283. https://doi.org/10.1017/s1355617712001282

Jaeger, T. F. (2010). Redundancy and reduction: speakers manage syntactic information density. *Cognitive Psychology*, *61*(1), 23–62.

Jaeger, T. F., & Buz, E. (2017). Signal Reduction and Linguistic Encoding. *The Handbook of Psycholinguistics*, pp. 38–81. https://doi.org/10.1002/9781118829516.ch3

Johnson, K. (1997). Speech perception without speaker normalization: An exemplar model. In K. Johnson & J. W. Mullennix (Eds.), *Talker variability in speech processing* (pp. 145–165). Academic Press.

Johnson, L. M., Di Paolo, M., & Bell, A. (2018). Forced Alignment for Understudied Language Varieties: Testing Prosodylab-Aligner with Tongan Data. *Language Documentation \& Conservation*, *12*, 80–123.

Jun, S.-A. (2014). Prosodic typology: by prominence type, word prosody, and macro-rhythm. *Prosodic Typology II*, pp. 520–539. https://doi.org/10.1093/acprof:oso/9780199567300.003.0017

Jurafsky, D., Bell, A., Gregory, M., & Raymond, W. D. (2001). Probabilistic relations between words. *Frequency and the Emergence of Linguistic Structure*, p. 229. https://doi.org/10.1075/tsl.45.13jur

Keuleers, E., Brysbaert, M., & Boris New. (2010). SUBTLEX-NL: A new measure for Dutch word frequency based on film subtitles. *Behavior Research Methods*, Vol. 42, pp. 643–650. https://doi.org/10.3758/brm.42.3.643

Kim, C., & Stern, R. M. (2008). Robust signal-to-noise ratio estimation basedon waveform amplitude distribution analysis. *Proceedings of Interspeech*, 2598–2601.

Kitamura, C. (2014). Hyper-Articulation of Child-Directed Speech. In P. J. Brooks & V. Kempe (Eds.), *Encyclopedia of language development* (pp. 273–274). Thousand Oaks, CA: SAGE Publications, Inc.

Kuhlen, A. K., & Brennan, S. E. (2013). Language in dialogue: when confederates might be hazardous to your data. *Psychonomic Bulletin & Review*, *20*(1), 54–72.

Kuperman, V., Pluymaekers, M., Ernestus, M., & Baayen, H. (2007). Morphological predictability and acoustic duration of interfixes in Dutch compounds. *The Journal of the Acoustical Society of America*, *121*(4), 2261–2271.

Kuznetsova, A., Brockhoff, P. B., & Christensen, R. H. B. (2017). lmerTest Package: Tests in Linear Mixed Effects Models. *Journal of Statistical Software*, Vol. 82. https://doi.org/10.18637/jss.v082.i13

Ladd, D. R. (1986). Intonational phrasing: the case for recursive prosodic structure. *Phonology*, *3*, 311–340.

Lam, T. Q., & Watson, D. G. (2010). Repetition is easy: Why repeated referents have reduced prominence. *Memory & Cognition*, *38*(8), 1137-1146.

Lee, Y.-C., Wang, B., Chen, S., Adda-Decker, M., Amelot, A., Nambu, S., & Liberman, M. (2015). A crosslinguistic study of prosodic focus. *2015 IEEE International Conference on Acoustics, Speech and Signal Processing (ICASSP)*. https://doi.org/10.1109/icassp.2015.7178873

Liberman, M. (2014, June 21). The shape of a spoken phrase in Mandarin. Retrieved April 5, 2019, from Language Log website: http://languagelog.ldc.upenn.edu/nll/?p=13101

Lieberman, P. (1960). Some Acoustic Correlates of Word Stress in American English. *The Journal of the Acoustical Society of America*, *32*(4), 451–454.

Liu, F., & Xu, Y. (2005). Parallel encoding of focus and interrogative meaning in Mandarin intonation. *Phonetica*, *62*(2-4), 70–87.

Liu, S., & Samuel, A. G. (2004). Perception of Mandarin Lexical Tones when F0 Information is Neutralized. *Language and Speech*, Vol. 47, pp. 109–138. https://doi.org/10.1177/00238309040470020101

Liu, Y., Fung, P., Yang, Y., Cieri, C., Huang, S., & Graff, D. (2006). HKUST/MTS: A Very Large Scale Mandarin Telephone Speech Corpus. *Chinese Spoken Language Processing*, pp. 724–735. https://doi.org/10.1007/11939993_73

Luke, S. G. (2017). Evaluating significance in linear mixed-effects models in R. *Behavior Research Methods*, *49*(4), 1494–1502.

Marslen-Wilson, W. D., & Welsh, A. (1978). Processing interactions and lexical access during word recognition in continuous speech. *Cognitive psychology*, 10(1), 29-63.

Maniwa, K., Jongman, A., & Wade, T. (2008). Perception of clear fricatives by normal-hearing and simulated hearing-impaired listeners. *The Journal of the Acoustical Society of America*, *123*(2), 1114–1125.

McAuliffe, M., Socolof, M., Mihuc, S., Wagner, M., & Sonderegger, M. (2017). Montreal Forced Aligner: Trainable Text-Speech Alignment Using Kaldi. *Interspeech 2017*. https://doi.org/10.21437/interspeech.2017-1386

McClelland, J. L., & Elman, J. L. (1986). The TRACE model of speech perception. *Cognitive psychology*, 18(1), 1-86.

McMillan, C. T., & Corley, M. (2010). Cascading influences on the production of speech: evidence from articulation. *Cognition*, *117*(3), 243–260.

Micci-Barreca, D. (2001). A preprocessing scheme for high-cardinality categorical attributes in classification and prediction problems. *ACM SIGKDD Explorations Newsletter*, *3*(1), 27.

National Institute of Standards and Technology. (n.d.). *Speech Quality Assurance (SPQA) Package* (Version 2.3). Retrieved from https://www.nist.gov/itl/iad/mig/tools

Nenkova, A., Brenier, J., Kothari, A., Calhoun, S., Whitton, L., Beaver, D., & Jurafsky, D. (2007). To Memorize or to Predict: Prominence Labeling in Conversational Speech. *Proceedings of NAACL-HLT*, 9–16.

Nomi, J. S., & Cleary, A. (2008). Recognition Memory for Proper versus Non-Proper Names. *PsycEXTRA Dataset*. https://doi.org/10.1037/e617962012-038

Nooteboom, S. G. (1981). Lexical retrieval from fragments of spoken words: Beginnings vs endings. *Journal of Phonetics*, *9*(4), 407-424.

Norris, D. (1994). Shortlist: A connectionist model of continuous speech recognition. *Cognition*, 52(3), 189-234.

Novak, J. R., Minematsu, N., & Hirose, K. (2016). Phonetisaurus: Exploring grapheme-to-phoneme conversion with joint n-gram models in the WFST framework. *Natural Language Engineering*, Vol. 22, pp. 907–938. https://doi.org/10.1017/s1351324915000315

Ouyang, I. C., & Kaiser, E. (2013). Prosody and information structure in a tone language: an investigation of Mandarin Chinese. *Language, Cognition and Neuroscience*, *30*(1-2), 57–72.

Peelle, J. E., & Davis, M. H. (2012). Neural Oscillations Carry Speech Rhythm through to Comprehension. *Frontiers in Psychology*, *3*, 320.

Pellegrino, F., Coupé, C., & Marsico, E. (2011). Across-Language Perspective on Speech Information Rate. *Language*, Vol. 87, pp. 539–558. https://doi.org/10.1353/lan.2011.0057

Peters, A., & Tse, H. (2016). *Evaluating the Efficacy of Prosody-lab Aligner for a Study of Vowel Variation in Cantonese*. Presented at the Workshop on Innovations in Cantonese Linguistics (WICL 3), The Ohio State University, Columbus, OH. Retrieved from http://d-scholarship.pitt.edu/27237/1/peters_tse_2016_03-12_wicl3_presentation.pdf

Piantadosi, S. T., Tily, H., & Gibson, E. (2011). Word lengths are optimized for efficient communication. *Proceedings of the National Academy of Sciences of the United States of America*, *108*(9), 3526–3529.

Pierrehumbert, J. B. (2001). Exemplar dynamics: Word frequency, lenition and contrast. In J. Bybee & P. Hopper (Eds.), *Frequency effects and the emergence of lexical structure* (pp. 137–157).

Pierrehumbert, J. B., & Beckman, M. (1988). *Japanese tone structure*. Mit Press.

Pluymaekers, M., Ernestus, M., & Harald Baayen, R. (2005a). Articulatory Planning Is Continuous and Sensitive to Informational Redundancy. *Phonetica*, Vol. 62, pp. 146–159. https://doi.org/10.1159/000090095

Pluymaekers, M., Ernestus, M., & Harald Baayen, R. (2005b). Lexical frequency and acoustic reduction in spoken Dutch. *The Journal of the Acoustical Society of America*, Vol. 118, pp. 2561–2569. https://doi.org/10.1121/1.2011150

Qun, L., Hua-Ping, Z., & Hao, Z. (2004). *计算所汉语词性标记集 (ICTPOS)* (Version 3.0). Retrieved from https://github.com/NLPIR-team/NLPIR/raw/master/NLPIR%20SDK/NLPIR-ICTCLAS/doc/ICTPOS3.0.doc

Raveh, E., Steiner, I., Siegert, I., Gessinger, I., & Möbius, B. (2019). Comparing phonetic changes in computer-directed and human-directed speech. In P. Birkholz & S. Stone (Eds.), *Elektronische Sprachsignalverarbeitung 2019: Tagungsband der 30. Konferenz Dresden* (pp. 42–49). TUDpress.

R Core Team. (2013). *R: A Language and Environment for Statistical Computing*. Retrieved from R Foundation for Statistical Computing website: http://www.R-project.org/

Robert Ladd, D. (2008). *Intonational Phonology*. Cambridge University Press.

Roettger, T. B., & Grice, M. (2019). The tune drives the text. *Language Dynamics and Change*, Vol. 9, pp. 265–298. https://doi.org/10.1163/22105832-00902006

Rooth, M. (1992). A theory of focus interpretation. *Natural Language Semantics*, Vol. 1, pp. 75–116. https://doi.org/10.1007/bf02342617

Rosa, E. C., Finch, K. H., Bergeson, M., & Arnold, J. E. (2015). The effects of addressee attention on prosodic prominence. *Language, Cognition and Neuroscience*, Vol. 30, pp. 48–56. https://doi.org/10.1080/01690965.2013.772213

Roten, T. (2017). PyNLPIR (Version 0.5.2). Retrieved from https://github.com/tsroten/pynlpir/commit/8d5e994796a2b5d513f7db8d76d7d24a85d531b1

Sano, S.-I. (2018). Durational contrast in gemination and informativity. *Linguistics Vanguard*, Vol. 4. https://doi.org/10.1515/lingvan-2017-0011

Schertz, J. (2013). Exaggeration of featural contrasts in clarifications of misheard speech in English. *Journal of Phonetics*, Vol. 41, pp. 249–263. https://doi.org/10.1016/j.wocn.2013.03.007

Schweitzer, K., Walsh, M., Calhoun, S., Schütze, H., Möbius, B., Schweitzer, A., & Dogil, G. (2015). Exploring the relationship between intonation and the lexicon: Evidence for lexicalised storage of intonation. *Speech Communication*, Vol. 66, pp. 65–81. https://doi.org/10.1016/j.specom.2014.09.006

Seyfarth, S. (2014). Word informativity influences acoustic duration: effects of contextual predictability on lexical representation. *Cognition*, *133*(1), 140–155.

Shaw, J. A. (2013). The phonetics of hyper-active feet: Effects of stress priming on speech planning and production. *Laboratory Phonology*, Vol. 4. https://doi.org/10.1515/lp-2013-0007

Shaw, J. A., Han, C., & Ma, Y. (2014). Surviving truncation: informativity at the interface of morphology and phonology. *Morphology, 24*(4), 407-432.

Shaw, J. A., & Kawahara, S. (2019). Effects of Surprisal and Entropy on Vowel Duration in Japanese. *Language and Speech*, *62*(1), 80–114.

Shaw, J. A., & Tyler, M. D. (2020). Effects of vowel coproduction on the timecourse of tone recognition. *The Journal of the Acoustical Society of America*, 147(4), 2511-2524.

Tang, K., & Bennett, R. (2018). Contextual predictability influences word and morpheme duration in a morphologically complex language (Kaqchikel Mayan). *The Journal of the Acoustical Society of America*, Vol. 144, pp. 997–1017. https://doi.org/10.1121/1.5046095

Tang, K., & Mandera, P. (In preparation). *SUBTLEX-CH 2.0*.

Tanner, J., Sonderegger, M., & Wagner, M. (2017). Production planning and coronal stop deletion in spontaneous speech. *Laboratory Phonology: Journal of the Association for Laboratory Phonology*, *8*(1), 15.

Torreira, F., & Ernestus, M. (2011). Vowel elision in casual French: The case of vowel /e/ in the word c'était. *Journal of Phonetics*, Vol. 39, pp. 50–58. https://doi.org/10.1016/j.wocn.2010.11.003

Torreira, F., & Ernestus, M. (2012). Weakening of intervocalic /s/ in the Nijmegen Corpus of Casual Spanish. *Phonetica*, *69*(3), 124–148.

Tremblay, A., & Tucker, B. V. (2011). The effects of N-gram probabilistic measures on the recognition and production of four-word sequences. *The Mental Lexicon*, 6(2), 302-324.

Tron, V. (2008). On the Durational Reduction of Repeated Mentions: Recency and Speaker Effects. In N. Calzolari, K. Choukri, B. Maegaard, J. Mariani, J. Odijk, S. Piperidis, & D. Tapias (Eds.), *Proceedings of the Sixth International Conference on Language Resources and Evaluation (LREC'08)* (pp. 2778–2780). Marrakech, Morocco: European Language Resources Association (ELRA).

Tseng, C.-Y., Pin, S.-H., Lee, Y., Wang, H.-M., & Chen, Y.-C. (2005). Fluent speech prosody: Framework and modeling. *Speech Communication*, Vol. 46, pp. 284–309. https://doi.org/10.1016/j.specom.2005.03.015

Turk, A. (2010). Does prosodic constituency signal relative predictability? A Smooth Signal Redundancy hypothesis. *Laboratory Phonology*, Vol. 1. https://doi.org/10.1515/labphon.2010.012

Turk, A. E., & Shattuck-Hufnagel, S. (2000). Word-boundary-related duration patterns in English. *Journal of Phonetics*, Vol. 28, pp. 397–440. https://doi.org/10.1006/jpho.2000.0123

Vainio, M., & Järvikivi, J. (2006). Tonal features, intensity, and word order in the perception of prominence. *Journal of Phonetics*, Vol. 34, pp. 319–342. https://doi.org/10.1016/j.wocn.2005.06.004

van Heuven, W. J. B., Mandera, P., Keuleers, E., & Brysbaert, M. (2014). Subtlex-UK: A New and Improved Word Frequency Database for British English. *Quarterly Journal of Experimental Psychology*, Vol. 67, pp. 1176–1190. https://doi.org/10.1080/17470218.2013.850521

Walker, A., & Hay, J. (2011). Congruence between 'word age'and 'voice age'facilitates lexical access. *Laboratory Phonology*, *2*(1), 219-237.

Watson, D. G., Arnold, J. E., & Tanenhaus, M. K. (2008). Tic Tac TOE: Effects of predictability and importance on acoustic prominence in language production. *Cognition*, Vol. 106, pp. 1548–1557. https://doi.org/10.1016/j.cognition.2007.06.009

Watson, D. G., Arnold, J. E., & Tanenhaus, M. K. (2010). Corrigendum to Tic Tac TOE: Effects of predictability and importance on acoustic prominence in language production. *Cognition*, Vol. 114, pp. 462–463. https://doi.org/10.1016/j.cognition.2010.01.007

Wedel, A. B. (2007). Feedback and regularity in the lexicon. *Phonology*, Vol. 24, pp. 147–185. https://doi.org/10.1017/s0952675707001145

Wedel, A., Nelson, N., & Sharp, R. (2018). The phonetic specificity of contrastive hyperarticulation in natural speech. *Journal of Memory and Language*, Vol. 100, pp. 61–88. https://doi.org/10.1016/j.jml.2018.01.001

Whalen, D. H., & Xu, Y. (1992). Information for Mandarin tones in the amplitude contour and in brief segments. *Phonetica*, *49*(1), 25–47.

Wilbanks, E. (2015). *The Development of FASE (Forced Alignment System for Espa\~nol) and implications for sociolinguistic research*. Presented at the New Ways of Analyzing Variation 44, Toronto, Canada. Retrieved from https://ericwilbanks.github.io/files/wilbanks_nwav_2015.pdf

Winter, B., & Christiansen, M. H. (2012). Robustness as a design feature of speech communication. *The Evolution of Language*. https://doi.org/10.1142/9789814401500_0050

Xu, Y., & Prom-on, S. (2019). Economy of Effort or Maximum Rate of Information? Exploring Basic Principles of Articulatory Dynamics. *Frontiers in Psychology*, Vol. 10. https://doi.org/10.3389/fpsyg.2019.02469

Xu, Y., & Wang, M. (2009). Organizing syllables into groups—Evidence from F0 and duration patterns in Mandarin. *Journal of Phonetics*, Vol. 37, pp. 502–520. https://doi.org/10.1016/j.wocn.2009.08.003

Yang, C. (2011). *The Acquisition of Mandarin Prosody by American Learners of Chinese as a Foreign Language (CFL)*. The Ohio State University.

Yang, C. (2016). *The Acquisition of L2 Mandarin Prosody: From experimental studies to pedagogical practice*. John Benjamins Publishing Company.

Yang, J., Zhang, Y., Li, A., & Xu, L. (2017). On the Duration of Mandarin Tones. *Interspeech 2017*, 1407–1411. ISCA: ISCA.

Yuan, J., Ryant, N., & Liberman, M. (2014). Automatic phonetic segmentation in Mandarin Chinese: Boundary models, glottal features and tone. *2014 IEEE International Conference on Acoustics, Speech and Signal Processing (ICASSP)*. https://doi.org/10.1109/icassp.2014.6854058

Zhang, H.-P. (2016). NLPIR-ICTCLAS (Version 2016). Retrieved from https://github.com/NLPIR-team/NLPIR/tree/master/NLPIR%20SDK/NLPIR-ICTCLAS

Zhang, H.-P., Yu, H.-K., Xiong, D.-Y., & Liu, Q. (2003). HHMM-based Chinese lexical analyzer ICTCLAS. *Proceedings of the Second SIGHAN Workshop on Chinese Language Processing -*. https://doi.org/10.3115/1119250.1119280

Zhao, Y., & Jurafsky, D. (2009). The effect of lexical frequency and Lombard reflex on tone hyperarticulation. *Journal of Phonetics*, Vol. 37, pp. 231–247. https://doi.org/10.1016/j.wocn.2009.03.002

Zipf, G. K. (1936). *The Psychobiology of Language*. London: Routledge.

Zipf, G. K. (1949). *Human Behaviour and the Principle of Least Effort*. Boston: Addison-Wesley Press.